\documentclass[10pt,twocolumn,letterpaper]{article}

\usepackage[accsupp]{axessibility}

\usepackage{bm}
\usepackage[abs]{overpic}
\AtBeginDocument{%
  \setlength\abovedisplayskip{1pt}
  \setlength\belowdisplayskip{1pt}}

\def\etal{et~al.\_}			  

\newcommand{\tb}{\textbf}

\def\sysname{DualVector\xspace}

\newlength\paramargin
\newlength\figmargin

\newlength\secmargin
\newlength\figcapmargin
\newlength\tabcapmargin

\newif\ifReview\Reviewfalse

\long\def\ignorethis#1{k}

\renewcommand{\tb}[1]{\textbf{#1}}

\newbox\jsavebox%

\makeatletter
\newcommand{\providelength}[1]{%
  \@ifundefined{\expandafter\@gobble\string#1}
   {
    \typeout{\string\providelength: making new length \string#1}%
    \newlength{#1}%
   }
   {
    \sdaau@checkforlength{#1}%
   }%
}

\usepackage{cvpr}              

\usepackage{graphicx}
\usepackage{amsmath}
\usepackage{amssymb}
\usepackage{booktabs}
\usepackage{multicol}
\usepackage{multirow}

\usepackage[pagebackref,breaklinks,colorlinks]{hyperref}

\makeatletter
\newcommand{\altaffiliation}[1]{%
  \let\@fnsymbol\@alph
  \footnotetext[#1]{#1}}
\makeatother

\usepackage{lipsum}

\newcommand\blfootnote[1]{%
  \begingroup
  \renewcommand\thefootnote{}\footnote{#1}%
  \addtocounter{footnote}{-1}%
  \endgroup
}

\usepackage[capitalize]{cleveref}
\crefname{section}{Sec.}{Secs.}
\Crefname{section}{Section}{Sections}
\Crefname{table}{Table}{Tables}
\crefname{table}{Tab.}{Tabs.}
\usepackage{bbding}
\usepackage[normalem]{ulem}

\def\cvprPaperID{407} 
\def\confName{CVPR}
\def\confYear{2023}
\begin{document}


\title{\sysname: Unsupervised Vector Font Synthesis with Dual-Part Representation}

\author{
Ying-Tian Liu$^{1}$\quad
Zhifei Zhang$^{2}$\quad
Yuan-Chen Guo$^{1}$\quad
Matthew Fisher$^{2}$\\
Zhaowen Wang$^{2}$\quad
Song-Hai Zhang$^{1*}$\\
$^{1}$ BNRist, Department of Computer Science and Technology, Tsinghua University\;
$^{2}$ Adobe Research\\
{\tt\small \{liuyingt20@mails., guoyc19@mails.,  shz@\}tsinghua.edu.cn}\\
{\tt\small\{zzhang, matfishe, zhawang\}@adobe.com}
}

\twocolumn[{%
\renewcommand\twocolumn[1][]{#1}%
\maketitle
\begin{center}
    \centering
    \vspace{-10mm}
    \captionsetup{type=figure}
    \includegraphics[width=\linewidth]{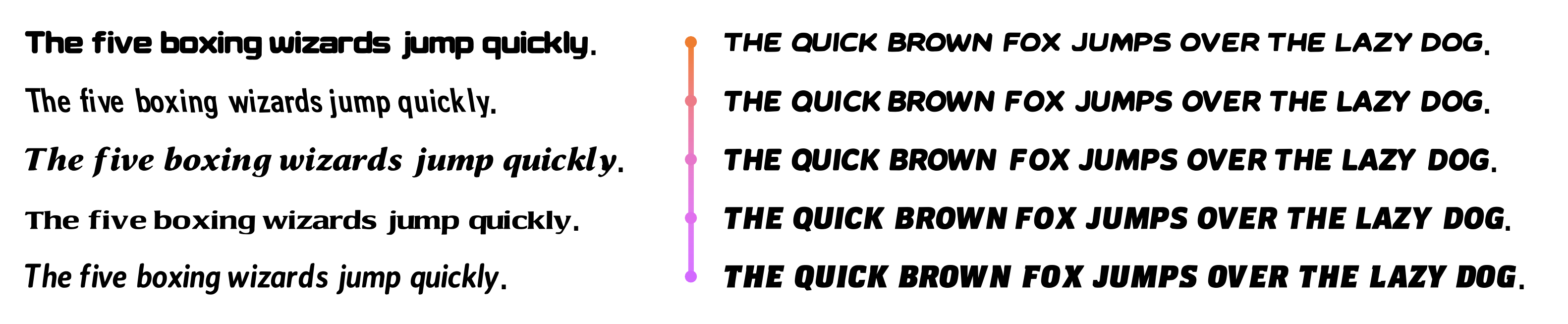}
    \vspace{-8mm}
    \captionof{figure}{High-quality vector fonts (copyable on electronic devices) synthesized by \sysname (left), with smoothly-interpolated styles (right). Both sentences are pangrams. Please zoom in for details.}
    \label{fig:teaser}
\end{center}%
}]
\vspace{3mm}

\begin{abstract}
   Automatic generation of fonts can be an important aid to typeface design. Many current approaches regard glyphs as pixelated images, which present artifacts when scaling and inevitable quality losses after vectorization. On the other hand, existing vector font synthesis methods either fail to represent the shape concisely or require vector supervision during training. To push the quality of vector font synthesis to the next level, we propose a novel dual-part representation for vector glyphs, where each glyph is modeled as a collection of closed ``positive'' and ``negative'' path pairs. The glyph contour is then obtained by boolean operations on these paths.  We first learn such a representation only from glyph images and devise a subsequent contour refinement step to align the contour with an image representation to further enhance details. Our method, named \sysname, outperforms state-of-the-art methods in vector font synthesis both quantitatively and qualitatively. Our synthesized vector fonts can be easily converted to common digital font formats like TrueType Font for practical use. The code is released at \href{https://github.com/thuliu-yt16/dualvector}{https://github.com/thuliu-yt16/dualvector}.
   
\end{abstract}


\section{Introduction}
Fonts with various styles play an important role in content display and distribution.
\blfootnote{$^*$Corresponding author}Excellent font design is time-consuming and labor-intensive. Recent machine learning innovations have made font generation possible, but how to automatically generate high-quality vector fonts remains a task of practical importance in the artistic and computer graphics and vision communities.

Benefiting from the development of image generation techniques, mainstream font synthesis methods~\cite{glyphgan, mcgan,font-cgan,tetgan,agisnet} could generate pixelated glyph images.
Despite the promising quality, images of glyphs incur aliasing artifacts on edges when discretely sampled,  and thus are not competent for high-quality rendering or printing at arbitrary resolutions.
To alleviate this problem, some methods~\cite{implicitfield,multi-implicit} adopt coordinate-based neural networks to model a glyph as a contiguous neural field, which have also shown great potential in modeling 3D geometry and scenes~\cite{implicitfield,deepsdf,nerf}. Although glyphs represented by the implicit field can be rendered at arbitrary resolutions, it is hard to preserve details in high-frequency regions such as edges and corners, not to mention the high computational costs as the network needs to be evaluated for every pixel. Researchers have made much effort to directly synthesize vector fonts~\cite{learned-svg,im2vec,deepsvg,deepvecfont} in recent years, with the main difficulty lying in finding a representation of vector graphics that can be encoded or decoded effectively in a deep learning framework. One typical approach represents a vector shape as a sequence of drawing commands and adopts sequence modeling techniques such as recurrent networks and transformers. The drawbacks are twofold: (1) Modeling command sequences can be much harder than images. There are infinitely many command sequences that correspond to the same-looking shape, which brings ambiguities in learning and makes it hard to construct an effective manifold for valid glyph shapes. (2) Ground-truth drawing commands are often required to provide sufficient supervision for high-quality modeling and synthesis. 

To overcome these challenges, we first propose a dual-part vector font representation, where each glyph shape is the union of a fixed number of \textit{dual parts}. Each \textit{dual part} is formed by the subtraction of a ``positive" and a ``negative" geometric primitive. While there are many choices for the geometric primitives~\cite{bspnet,cvxnet,learning-implicit-glyph}, we adopt closed B\'ezier paths for their great representational ability. They are also widely supported in digital font formats which makes it easy to convert our representation to these formats for practical use. We reduce the problem of predicting complicated drawing command sequences to predicting multiple basic primitives. From this perspective, both manifold learning and latent space interpolation become more feasible.

Based on the dual-part representation, we introduce \sysname, a method to learn such a representation for high-quality font modeling and synthesis from only glyph images without any vector supervision. A straightforward way to achieve this is to directly optimize the parameters of the B\'ezier curves with differentiable rendering techniques for vector graphics~\cite{diffvg}. However, this approach easily gets stuck in the local minima as valuable gradients are only defined at shape intersections. Taking inspiration from implicit field training for 2D and 3D shapes~\cite{bspnet,deepsdf,learning-implicit-glyph}, we supervise the occupancy value derived from the analytical expression of the B\'ezier curves and adopt an initialization strategy based on unsigned distance field (UDF) to provide dense gradients across the entire pixel space. For local detail fidelity, we also train a glyph image generation model and devise a subsequent contour refinement step to align the contour of the vector shape with that of the image by differentiable rendering~\cite{diffvg}.
We compare our approach with state-of-the-art methods in font modeling and generation and demonstrate the superior quality of our vector font outputs. Our main contributions are:

\begin{itemize}
    \item A new dual-part font representation based on boolean operations of B\'ezier paths, which enables efficient shape modeling and unsupervised  manifold learning.


    \item A method named \sysname that models both the dual-part and pixelated representation, and introduces a contour refinement step to obtain vector fonts with richer details as well as a UDF initialization strategy for better convergence.
    
    
    
    \item \sysname achieves state-of-the-art quality in font modeling and generation, with outputs that can be easily converted to common digital font formats.
    

\end{itemize}

\section{Related Work}

\subsection{Glyph Image Generation}

Benefiting from the development of image generation\cite{vae14, goodfellow14,pix2pix}, either black-white~\cite{dcfont,zi2zi,glyphgan, jointfontgan} or artistic glyph image generation\cite{tetgan,mcgan,agisnet, dsenet} is well explored in the past decade. 
MC-GAN\cite{mcgan} synthesized ornamented glyphs for capital letters in an end-to-end manner from a small subset of the same style. 
Attr2Font\cite{attribute2font} generated visually pleasing results according to style attributes and content references. Most of these methods follow the disentanglement of style and content codes, and so does the image generation branch in \sysname.
But these glyph image generation efforts do not step outside the 2D pixelated representation which sets an upper bound for rendering quality, efficiency, and practicality. 


\subsection{Image Rasterization and Vectorization}
Rasterization and vectorization are dual problems in image processing.  
Previous efforts on rasterization have typically focused on anti-aliasing\cite{fabris1997, green2007improved, randomaccess} and high-performance rendering on modern graphics hardware\cite{loop2005resolution, yoshi2006,vineet2015,li2016gpu}. Traditional image vectorization approaches rely on region segmentation, line extraction, and contour tracing for gray-scale images\cite{potrace}, pixel arts\cite{depixeling}, animations\cite{cart_ani} and general images\cite{grad_mesh, ardeco,vectormagic,adobetrace, inv-diff-curves, tcb-spline}. 
Recently, researchers have turned to bridging the raster and vector domains~\cite{diffvg,live}, enabling existing well-established algorithms for raster images to be also applied to vector representations.
DiffVG \cite{diffvg} is a differentiable rendering technique for vector graphics  enabling the optimization of vector primitive parameters based on raster criteria. 
In our work, it is employed to align the vector and pixel representation in the contour refinement step. 

\begin{figure*}[t]
\centering
\scalebox{0.9}{
\begin{overpic}[scale=0.55]{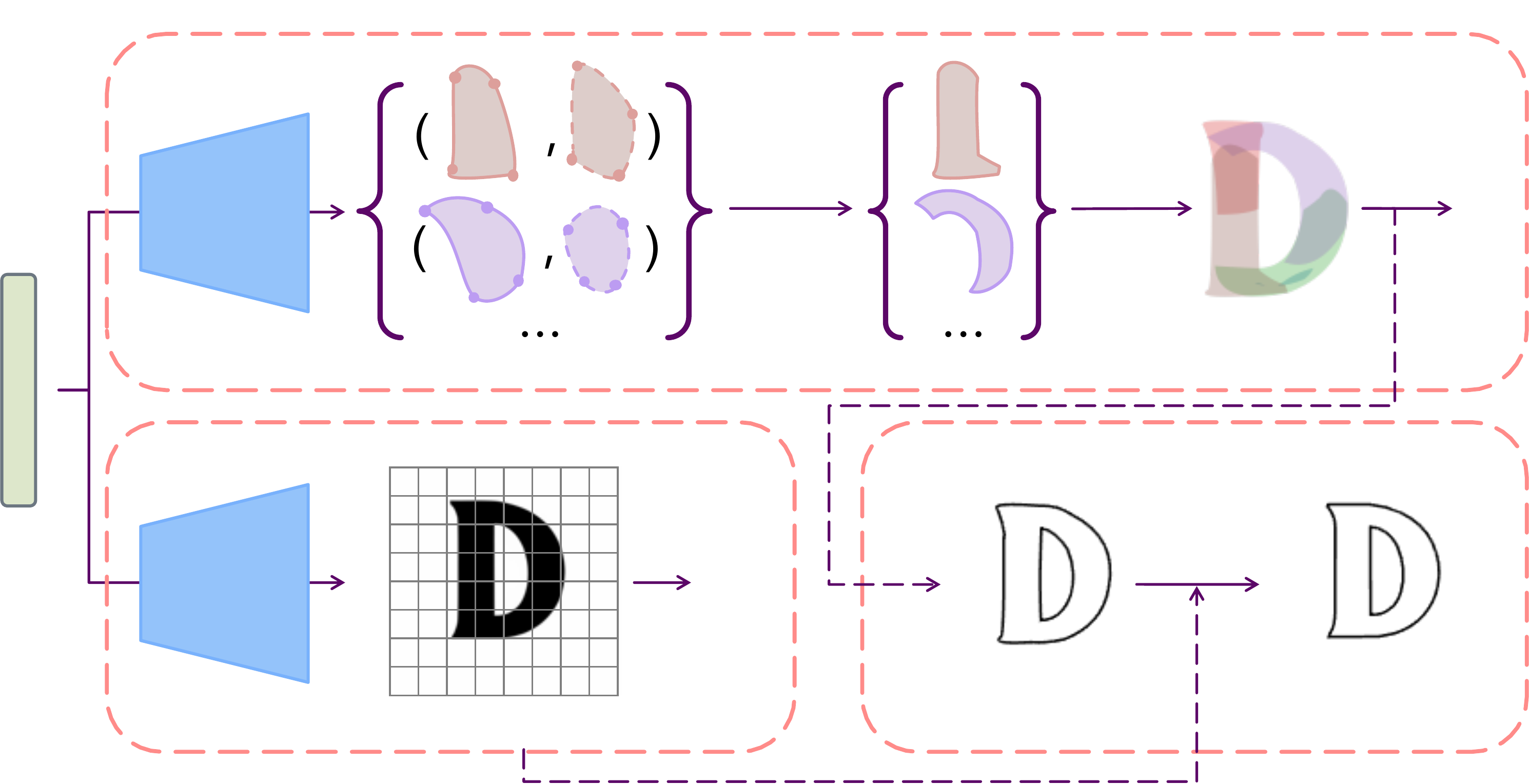}
\put(2.8,75){$z$}
\put(63,175){\scalebox{1.5}{$\mathcal{D}_P$}}
\put(63,59){\scalebox{1.5}{$\mathcal{D}_I$}}
\put(115,15){Reconstructed Image $\mathcal{I}$}
\put(220,60){$\mathcal{L}_I$}
\put(455,175){$\mathcal{L}_P$}
\put(295,25){Initialization $\partial{\mathcal{O}}$}
\put(400,25){Vector Output}
\put(115,126){Dual Parts  $\{(P_i, Q_i)\}_{i=1}^N$}
\put(272,126){$\{P_i-Q_i\}_{i=1}^N$}
\put(385,126){$\mathcal{O}/\hat{\mathcal{O}}$}
\put(354,65){$\mathcal{L}_{refine}$}
\put(42, 215){(a) \textit{Vector Branch}}
\put(274, 96){(c) \textit{Contour Refinement}}
\put(42, 96){(b) \textit{Image Branch}}
\put(229,180){\small Subtract}
\put(337,180){\small Union}
\put(145,200){$P_1$}
\put(181,200){$Q_1$}
\put(145,162){$P_2$}
\put(181,162){$Q_2$}
\end{overpic}}
\vspace{-2mm}
\caption{\sysname contains the following components: (a) a vector branch that maps the latent code $z$ to several closed B\'ezier paths which are further gathered to form a global shape of the glyph; (b) an image branch that generates pixelated images with faithful details; (c) a contour refinement step that obtains the contour via boolean operations on dual-parts and optimizes it with the image guidance at inference time. Both the vector and the image branch are trained using the glyph images. Dual parts are distinguished by their colors.}
\label{fig:method}
\vspace{-4mm}
\end{figure*}
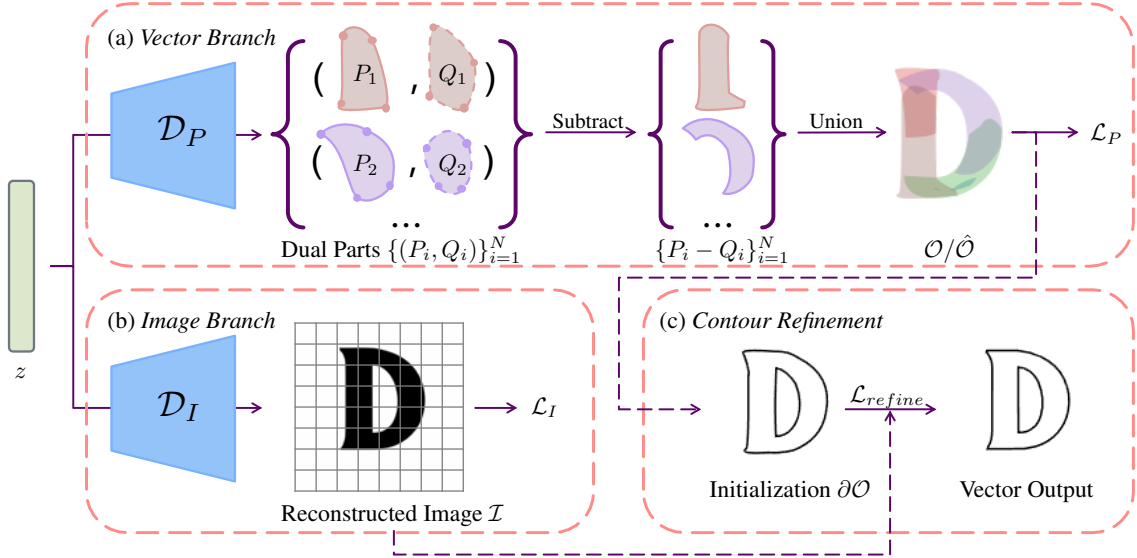

\subsection{Vector Font Generation}
SVG-VAE~\cite{learned-svg} was the first attempt to build a sequential generative model on SVG fonts. DeepSVG~\cite{deepsvg} developed a hierarchical transformer-based generative model for complex SVG icons generation and interpolation.
Exploiting implicit neural representation, multi-implicits~\cite{multi-implicit} modeled fonts with a permutation-invariant set of learned implicit functions which can preserve font features, such as smooth edges and sharp corners, and generate good interpolation results. Liu \etal~\cite{learning-implicit-glyph} proposed a primitive-based representation, which views glyph shapes as the combination of primitives enclosed by implicit curves. 
Implicit representation-based methods can convert the zero-level set of their output fields to contour representation with the 2D marching cube~\cite{marchingcube} algorithm, although it produces multiple segmented lines and lacks accuracy and efficiency in rendering and editing.  
Im2Vec\cite{im2vec} can synthesize complex vector graphics with varying topologies without vector supervision. It rendered layered B\'ezier paths predicted from image features using DiffVG\cite{diffvg} and trains the system in an auto-encoding manner.
DeepVecFont\cite{deepvecfont} adopted a dual-modality learning framework that utilizes both sequence and image features for high-quality vector font synthesis. But the results it produces may have the wrong topology, including self-intersection or invalid paths. In contrast to these methods, our approach does not require vector supervision and is very efficient in vector representation, with high quality glyph generation.



\section{\sysname}
In this section, we first illustrate the proposed dual-part representation for fonts (\cref{sec:dual-part}) and then introduce the components of \sysname for training this representation from glyph images (\cref{sec:vector-branch,sec:image-branch,sec:refinement}) and how the model can be used for font reconstruction and generation (\cref{sec:implementation}). In \sysname, we adopt a joint representation of vector and image for fonts, as shown in \cref{fig:method}. The latent space is shared by both modalities, and we associate each glyph with a latent code $z \in \mathbf{R}^d$ produced by the task-specific encoding process. The latent code $z$ is fed into the two decoding branches to obtain a dual-part representation and an image representation respectively. The output of the vector branch is further processed by a contour refinement step to generate the final contour representation under the image guidance.


\subsection{Dual-Part Representation} \label{sec:dual-part}
In our dual-part representation, we consider the closed parametric B\'ezier paths as our basic primitives for several reasons: (1) B\'ezier paths are powerful enough to approximate various shapes; (2) The contour of a closed shape produced by boolean operations on B\'ezier paths is still a set of B\'ezier paths, i.e., the closure property;
(3) B\'ezier paths are widely used in modern font design, making the learned representation highly applicable. For convenience, we use a triplet of control points $(\bm{a},\bm{b},\bm{c})$ to denote a quadratic B\'ezier curve $B(\cdot; \bm{a}, \bm{b}, \bm{c})$ determined by the control points:\begin{equation}
    B(t;\bm{a},\bm{b},\bm{c}) = (1-t)^2\bm{a} + (1-t)t\bm{b}+t^2\bm{c}t\in [0,1]
\end{equation}
Since a glyph often consists of a number of strokes with similar appearances, we propose to represent each glyph $g$ with $N$ closed B\'ezier paths $\{P_i\}_{i=1}^N$. 
Each path $P_i$ is defined by $M$ end-to-end connected quadratic B\'ezier curves $\{B_{i,j}(\bm{x}_{i,2j-1}, \bm{x}_{i,2j}, \bm{x}_{i,2j+1})\}_{j=1}^M$ ($\bm{x}_{i, 2M+1} = \bm{x}_{i, 1}$). We introduce the occupancy field of a closed path $P_i$ as 
\begin{equation}
    \mathcal{O}_{P_i}(\bm{x}) = \left\{\begin{array}{cc}
       1  ,& \bm{x} \in P_i \\
       0  ,& \bm{x} \notin P_i
    \end{array}\right.
\end{equation}
The entire glyph shape $g$ could simply be denoted as the union of all the paths $\cup_i P_i$. We normalize the coordinate range of the glyph to $[-1,1]^2$. Thus given any point $\bm{x}$ on the canvas $[-1,1]^2$, we can determine whether it is inside the represented glyph shape by a maximum operation, leading to the occupancy field of $g$:
\begin{equation}
    \mathcal{O}(\bm{x}) = \max_i \mathcal{O}_{P_i}(\bm{x}) = \left\{\begin{array}{cc}
       1  ,& \bm{x} \in \cup_i P_i \\
       0  ,& \bm{x} \notin \cup_i P_i
    \end{array}\right. 
\end{equation}

However, we find that the above representation can model fonts with convex curves on the contour but struggles to reconstruct shapes with holes. Therefore, in practice, we pair each path $P_i$ with a negative path $Q_i$, and term this pair $(P_i, Q_i)$ as a ``dual part". 
In this way, $g$ is represented as $\cup_i (P_i - Q_i)$. The occupancy field $\mathcal{O}$ is then derived as:
\begin{equation}
    \mathcal{O}(\bm{x}) = \max_i [\min(\mathcal{O}_{P_i}(\bm{x}), 1-\mathcal{O}_{Q_i}(\bm{x}))] = \left\{\begin{array}{cc}
       1  ,& \bm{x} \in g \\
       0  ,& \bm{x} \notin g
    \end{array}\right. 
\end{equation}

Even though $\mathcal{O}$ is the analytical occupancy field of $g$, it is not differentiable w.r.t. the parameters of the paths. Therefore, in order to apply gradient descent to learn such a representation, we calculate the approximate occupancy field $\hat{\mathcal{O}}$ from the signed distance field (SDF) of $g$. Since the distance $d(p; B)$ from any point $p$ to a B\'ezier curve $B$ can be derived analytically, the SDF of a path can be calculated differentiably w.r.t. its control points as follows:
\begin{equation}
    s_{P_i}(\bm{x}) = [2\mathcal{O}_{P_i}(\bm{x}) - 1]\min_j d(\bm{x};B_{i,j})
\end{equation}
Following previous vector graphics rasterization techniques\cite{randomaccess,diffvg,multi-implicit}, we approximate $\mathcal{O}$ by analytical pixel pre-filtering with a parabolic kernel $\alpha$:
\begin{equation}
\begin{aligned}
    & \hat{\mathcal{O}}_{P_i}(\bm{x}) = \alpha(s_{P_i}(\bm{x}))\\
    & \hat{\mathcal{O}}= \max_i [\min(\hat{\mathcal{O}}_{P_i}, 1-\hat{\mathcal{O}}_{Q_i})] 
\end{aligned}
\end{equation}


\subsection{Vector Branch}
\label{sec:vector-branch}

The vector branch takes a latent code $z$ and outputs a set of dual-parts that represent a vector glyph.
The path decoder $\mathcal{D}_P$ directly predicts the control points of the positive paths $\{P_i\}_{i=1}^N$ and the negative ones $\{Q_i\}_{i=1}^N$ from $z$:
\begin{equation}
    \{\bm{x}_{i, j}\} = \mathcal{D}_P(z), 1 \le i \le 2N, 1 \le j \le 2M
\end{equation}
where $\{\bm{x}_{i, j}\}_{i=1}^{N}$ are control points for $\{P_i\}_{i=1}^N$ and $\{\bm{x}_{i, j}\}_{i=N+1}^{2N}$ are control points for $\{Q_i\}_{i=1}^N$.


\subsection{Image Branch}
\label{sec:image-branch}
To combine with the advantages of image generation approaches, we train an auxiliary image branch that maps $z$ to a pixelated image $\mathcal{I}$ of the glyph it represents. Here we adopt a CNN-based decoder $\mathcal{D}_I$:
\begin{equation}
    \mathcal{I} = \mathcal{D}_I(z)
\end{equation}
where $\mathcal{I}$ is a gray-scale image in shape $H\times W$. It provides detailed shape guidance to refine the vector shape contour. 

\subsection{Contour Refinement}
\label{sec:refinement}
Although the dual-part representation produced by the vector branch can already be rendered, there still exists a gap between it and modern contour representations for fonts. Therefore, we first convert it to the contour of the glyph $\partial \mathcal{O}$ with an off-the-shelf tool Paper.js\cite{paperjs} which can perform arbitrary boolean operations on SVG paths. 
Here the contour $\partial \mathcal{O}$ is a set of K closed B\'ezier paths.
\begin{equation}
    \partial \mathcal{O} = \{C_1, C_2, ..., C_K\} 
\end{equation}
where $K$ denotes the number of independent paths. Each individual part or hole increases $K$ by 1. $C_i$ is a closed B\'ezier path composed of $l_i$ quadratic B\'ezier segments $\{\hat{B}_{i,j}(\bm{\hat{x}}_{i,2j-1}, \bm{\hat{x}}_{i,2j}, \bm{\hat{x}}_{i,2j+1})\}_{j=1}^{l_i}$. 

After this conversion, we exploit the image generation results to further refine $\partial \mathcal{O}$ with differentiable rendering techniques~\cite{diffvg}. $\mathcal{I}$ could naturally serve as a pixelated version of the occupancy field $\mathcal{O}$. During the refinement, we devise several strategies to improve the quality and efficiency of our representation:
\paragraph{Subdivision}
To improve the representational ability, we split long B\'ezier curves (with a length greater than 10\% of the canvas width) at their midpoints. 
\paragraph{Simplification}
To improve the efficiency of the contour representation, we replace the B\'ezier curves with a line segment connecting its two endpoints if they are close enough. For a B\'ezier curve $(\bm{a}, \bm{b}, \bm{c})$, we simply define how faithfully the segment $\bm{ac}$ could represent the curve by the angle between $\vec{\bm{ba}}$ and $\vec{\bm{bc}}$. Also, we examine if there exist two adjacent B\'ezier segments that could be combined as a single one. Since all quadratic B\'ezier curves are parts of a parabola curve, we derive the parameters of the corresponding parabola curves from the control points and combine the two curves if their parameters are close enough. We also aggregate the endpoints of lines and curves that are too short.
\paragraph{Pruning}
Prior to refinement, some B\'ezier paths on the contour may enclose extremely small regions which will confuse the refinement process and produce redundant curves that eventually converge to a single point. We trim those closed paths $C_i$ with areas under a certain threshold.

More implementation details of these strategies can be found in the supplemental material.

\subsection{Implementation and Training} \label{sec:implementation}

In this subsection, we will dive into how we utilize the proposed representation in both the font reconstruction and generation tasks.

\subsubsection{Optimization Objectives}
Optimization objectives can be divided into those at training time and those at inference time. The training-time objectives are task-related. 

\paragraph{Font reconstruction.}
For the font reconstruction task, the input to our system is a gray-scale image of a glyph and the output is the corresponding vector representation. The pixelated image $\mathcal{I}_{g}$ of the glyph $g$ is first passed through an image encoder $\mathcal{E}$ to obtain the latent code $z$:
\begin{equation}
    z = \mathcal{E}(\mathcal{I}_g)
\end{equation}

The latent code is then decoded by our vector branch and image branch into dual-parts $\{(P_i, Q_i)\}_{i=1}^N$ and a pixelated output $\mathcal{I}$. We optimize the trainable modules $\mathcal{E}$, $\mathcal{D}_I$ and $\mathcal{D}_P$ simultaneously with the following loss functions:
\begin{equation}
    \mathcal{L} = \lambda_P\mathcal{L}_P + \lambda_I\mathcal{L}_I
    \label{loss:rec}
\end{equation}
where we adopt an extra perceptual loss~\cite{lpips}:
\begin{equation}
\begin{aligned}
\mathcal{L}_P & = \mathbb{E}_{\bm{x}\sim[-1,1]^2}[||\hat{\mathcal{O}}(\bm{x}) - \mathcal{I}_g(\bm{x})||_1] \\
\mathcal{L}_I & = ||\mathcal{I} - \mathcal{I}_g||_2 + \text{LPIPS}(\mathcal{I}, \mathcal{I}_g)
\end{aligned}
\end{equation}
\vspace{-5mm}
\paragraph{Font generation.}
The input to the font generation task is multiple style reference images $\{R_i\}_{i=1}^{N_R}$ along with the corresponding character label $T$. The number of references $N_R$ may vary during training to allow the model to accept a variable number of references during inference. The style references are first mapped to their features $\{f_i\}_{i=1}^{N_R}$ with an image encoder $\mathcal{E}_f$. 
We then apply stacked self-attention layers (SA) to aggregate these features to a style latent code $z_{style}$ with a variational loss to enable font sampling.
Character labels are mapped to vector representations by a learnable label embedding layer and then fused with $z_{style}$ to yield the latent code $z_T$ of the target glyph. Then $z_T$ is fed to the decoders to generate the glyph, the same as the reconstruction task. The whole process can be described as follows: 
\begin{equation}
\begin{aligned}
    f_i & = \mathcal{E}_f(R_i) \\
    \mu, \sigma & = \text{SA}(f_1, f_2, ..., f_{N_R}) \\
    z_{style} & = \mu + \epsilon\sigma, \epsilon\sim\mathcal{N}(0,1) \\
    z_T & = \text{FFN}([z_{style}, T])
\end{aligned}
\end{equation}
To ease training, we use the pre-trained $\mathcal{E}$ from the reconstruction task to encode ground truth output $g_T$ and train the encoder with the following guidance:
\begin{equation}
    \mathcal{L}_{latent} = ||z_T - \mathcal{E}(g_T)||_2
\end{equation}
To enable font sampling, we adopt a Kullback-Leibler divergence loss\cite{vae14}:
\begin{equation}
    \mathcal{L}_{kl} = \text{KL}(\mathcal{N}(\mu, \sigma^2) || \mathcal{N}(0,1))
\end{equation}
Next, the encoder is fine-tuned along with the pre-trained $\mathcal{D}_I$ and $\mathcal{D}_P$ from the reconstruction task using the loss in \ref{loss:rec} with $\mathcal{L}_{kl}$. 

\vspace{-4mm}
\paragraph{Refinement}
Both the reconstruction and generation tasks share the same contour refinement process. Given $\partial{\mathcal{O}}$ and $\mathcal{I}$, the refinement process produces the optimized glyph contour in SVG format by inference-time optimization. 
In the optimization, we utilize DiffVG\cite{diffvg}, denoted as DR, to render the contour in a differentiable manner. The control points of $\mathcal{O}$ are refined to minimize the following loss function:
\begin{equation}
    \mathcal{L}_{refine} =  \mathcal{L}_{ras} + \lambda_{reg}\mathcal{L}_{reg} 
\end{equation}
where $\mathcal{L}_{render}$ is the photometric loss between the rasterized image and the image branch output. 
\begin{equation}
    \mathcal{L}_{ras} =  ||\text{DR}(\partial{\mathcal{O}}) - \mathcal{I}||_1
\end{equation}
The regularization term limits the overall length of $\mathcal{O}$. 
\begin{equation}
     \mathcal{L}_{reg} =  \sum_{i=1}^K \text{len}(C_i)
\end{equation}

\subsubsection{Unsigned Distance Field Warm-up}
Although the vector branch is end-to-end trainable, the gradients only appear near the boundaries. Therefore, to overcome the vanishing gradient problem, we add an extra loss based on the unsigned distance field (UDF) of the vector parts to warm up the system training. 

In general, for two closed curves $c_1, c_2$, we cannot obtain the exact SDF of the shape after arbitrary boolean operations on the curves by point-wise operations on their respective SDFs. For the union of two SDFs, as an approximation, we have 
\begin{equation}
    s_{c_1 \cup c_2}(\bm{x}) \left\{\begin{array}{cc}
       = \min (s_{c_1}(\bm{x}), s_{c_2}(\bm{x})) ,& \bm{x} \notin c_1 \cup c_2 \\
        \le \min (s_{c_1}(\bm{x}), s_{c_2}(\bm{x})) ,& \bm{x} \in c_1 \cup c_2
    \end{array}\right.
\end{equation}
The $\min(\cdot)$ operation gives a correct SDF outside the shape but an inaccurate SDF inside the shape. Similarly, the $\max(\cdot)$ gives a correct SDF inside and an inaccurate SDF outside for the intersection of two SDFs. Therefore, it is hard to calculate the accurate SDF $s_{\mathcal{O}}$ for the  glyph. Fortunately, we only aim to provide a coarse initialization for the dual parts, preventing the optimization process from falling into local minima early in the training. We find that using an approximate unsigned distance field works well to achieve this goal. Here we define the UDF $u$ as 
\begin{equation}
    u(\bm{x}) = \max(0, s(\bm{x}))
\end{equation}
where $\min(\cdot)$ gives a correct global UDF for unions of shapes. The approximate UDF $\hat{u}_g$ is computed as 
\begin{equation}
    \hat{u}_g = \min_i [\max(u_{P_i}, u_{Q_i})]
\end{equation}
where 
\begin{equation}
\begin{aligned}
    u_{P_i}(\bm{x}) & = \max(s_{P_i}(\bm{x}), 0) \\
    u_{Q_i}(\bm{x}) & = \max(-s_{Q_i}(\bm{x}), 0) 
\end{aligned}
\vspace{-1mm}
\end{equation}
We apply the following losses to warm up the training process for the font reconstruction task:
\begin{equation}
\begin{aligned}
    \mathcal{L} & = \lambda_P\mathcal{L}_P + \lambda_I\mathcal{L}_I + \lambda_{u}\mathcal{L}_u\\
    \mathcal{L}_u & = \mathbb{E}_{\bm{x}\sim[-1,1]^2}[||\hat{u}_g(\bm{x}) - u_g(\bm{x})||_1]
\end{aligned}
\end{equation}
where the ground truth UDF $u_g$ is derived from the pixel image $g$. 





\section{Experiments}

\subsection{Dataset}
We use the SVG-Fonts dataset from SVG-VAE\cite{learned-svg} for all the experiments. Following DeepVecFont\cite{deepvecfont}, we sample a subset of the dataset with 8035 fonts for training and 1425 fonts for evaluation. The resolution of input glyph images is set to $128\times128$. 
Please refer to the supplementary material for implementation details.


\begin{figure}[th]
    \centering
    \setlength{\belowcaptionskip}{-20pt}
    
    \scalebox{0.9}{
    \begin{overpic}[scale=0.30]{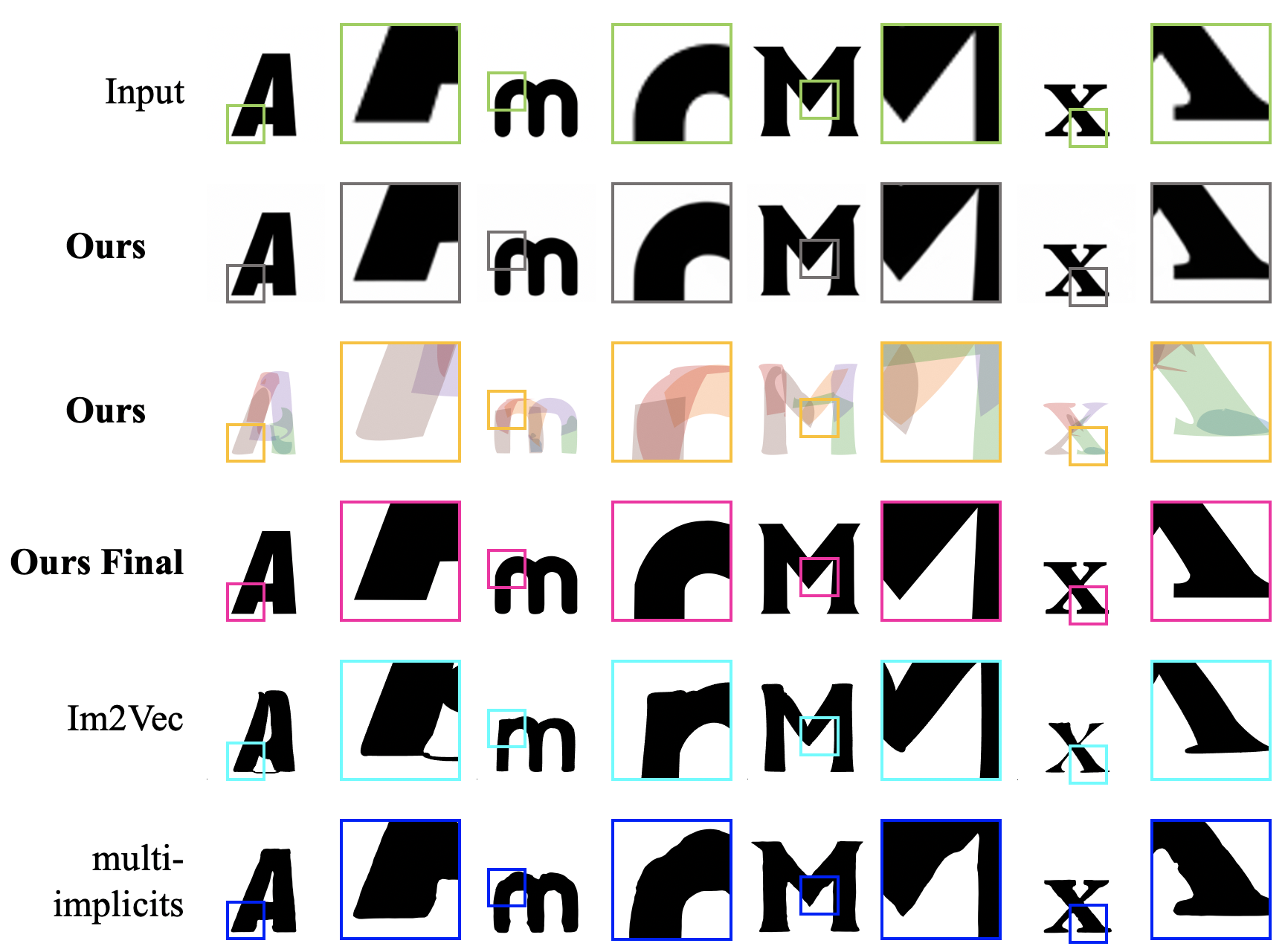}
    \put(32,107.5){$\mathcal{O}$}
    \put(32,140){$\mathcal{I}$}
    \end{overpic}}
    \caption{Comparison of font reconstruction quality. We show the input glyph image row by row, our reconstructed image $\mathcal{I}$ from the image branch, dual-parts $\mathcal{O}$ from the vector branch, and our final result after contour refinement, as well as vector outputs from alternative methods.}
    \label{fig:recon}
\end{figure}

\begin{figure*}[t]
\centering
\setlength{\belowcaptionskip}{-15pt}
\includegraphics[width=\linewidth]{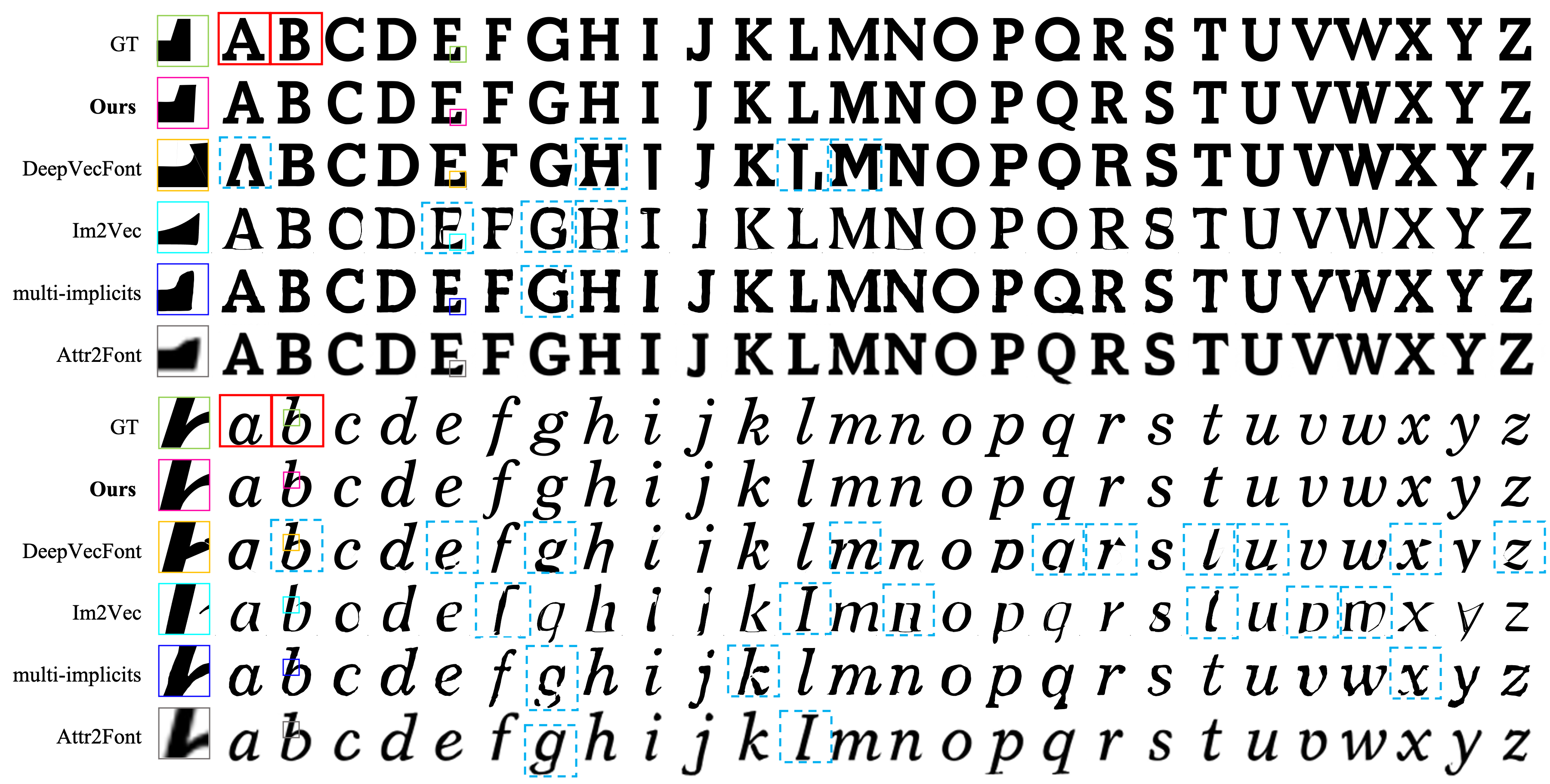}
\vspace{-8mm}
\caption{Font generation results of Attr2Font, Im2Vec, Multi-Implicits, DeepVecFont, and \sysname. The input style references are marked with red boxes. We trace the zero-level sets of SDFs in a piece-wise linear way to get SVGs from Multi-Implicits\cite{multi-implicit}. Typical failure cases are marked in blue dashed boxes. Some details are zoomed-in in the leftmost column. }
\label{fig:comple}
\end{figure*}

\subsection{Font Reconstruction}

In this experiment, we investigate how faithfully different methods can reconstruct the input glyph images with vector outputs, which can also be regarded as font vectorization. Im2Vec~\cite{im2vec} and our method directly produce vector representations from glyph images, while for Multi-Implicits\cite{multi-implicit}, we adopt the 2D marching cube algorithm and extract the contours. As can be seen from a qualitative comparison in \cref{fig:recon}, our representation could faithfully reconstruct global shapes as well as local details from the pixelated input. Im2Vec\cite{im2vec} could only preserve coarse shapes and suffers from wrong topologies as shown in the ``A" example as it is easy to fall into local optimum  without the global gradient.
Multi-Implicits\cite{multi-implicit} results in finer details than Im2Vec\cite{im2vec} but still 
exhibit unsmooth edges and rounded corners. To quantitatively evaluate how accurate the vector outputs are, we calculate three metrics, namely SSIM, L1, and s-IoU, between the rendered images and the ground truth images under different resolutions. \cref{tab:reconstruction} shows that 
our method surpasses the two alternatives by a clear margin on all metrics. Using parametric curves as primitives, our method, along with Im2Vec~\cite{im2vec}, maintains a stable L1 error across all resolutions for the smooth edges, while for Multi-Implicits\cite{multi-implicit} the L1 error gets larger as the resolution increases, resulting from the high-frequency noises at the extracted boundaries.

\begin{table}%
\caption{Quantitative comparison with Im2Vec~\cite{im2vec} and Multi-Implicits~\cite{multi-implicit} on three image-level metrics at different resolutions for the font reconstruction task. The gray scale is normalized to [0,1]. s-IoU, from \cite{multi-implicit}, measures the overlap.}
\vspace{-1mm}
\label{tab:reconstruction}
\centering
\begin{tabular}{rcccc}
  \toprule
  Method & Resolution & SSIM$\uparrow$ & L1$\downarrow$ & s-IoU$\uparrow$ \\
  \midrule
  \multirow{3}*{\textbf{Ours}} & $128^2$ & \tb{0.9436}  & \tb{0.0137} & \tb{0.8879} \\
  & $256^2$ & \tb{0.9483}  & \tb{0.0136} & \tb{0.9068} \\
  & $512^2$ &  \tb{0.9579} & \tb{0.0136} & \tb{0.9135} \\
  \midrule
  
  \multirow{3}*{Multi-Implicits} & $128^2$ & 0.9231 & 0.0183 & 0.8709 \\
  & $256^2$ & 0.9262 & 0.0208 & 0.8721 \\
  & $512^2$ & 0.9419 & 0.0213 & 0.8739 \\
  \midrule
  
  \multirow{3}*{Im2Vec} & $128^2$  & 0.7800 & 0.0504 & 0.6832 \\
  & $256^2$ & 0.8566 & 0.0504 & 0.6847 \\
  & $512^2$ & 0.8957 & 0.0504 & 0.6851 \\
  
  \bottomrule
\end{tabular}
\vspace{-5mm}
\end{table}

\subsection{Font Generation}
To evaluate the generation ability of our method, we adopt the few-shot font generation setting, where a subset of glyphs with the same style is given and models are trained to produce a complete set of glyphs that are consistent with the input glyphs in style. This task is also dubbed ``font complement" or ``font style transfer" in other works. We compare our method with several popular methods in font generation: DeepVecFont\cite{deepvecfont}, Multi-Implicits\cite{multi-implicit}, Im2Vec\cite{im2vec} and Attr2Font\cite{attribute2font}. For a fair comparison, we use the same four characters, `A', `B', `a', and `b' as style references for each font and evaluate image-level metrics on the generated characters. Since Im2Vec\cite{im2vec} is an image vectorization method, we apply it to the images generated by Attr2Font\cite{attribute2font}. Multi-Implicits\cite{multi-implicit} follows an auto-decoder architecture, so we freeze the decoder and find the optimal latent vector for a font by minimizing the losses between the given style references and the predicted ones using gradient descent.

As illustrated in \cref{fig:comple}, DeepVecFont\cite{deepvecfont} may generate characters with wrong topologies due to the incorrect initial contour prediction, such as `A', `L', and `Z'. The fixed topology in the optimization process makes it difficult to produce the necessary details, which are particularly symbolic in serif fonts. 
In addition, unconstrained contour generation may also produce self-intersections, like the `g' and `r', which are not easy to fix in subsequent processing. Im2Vec\cite{im2vec} tends to produce rounded corners and is only able to produce the coarse style, lacking details. Multi-Implicits\cite{multi-implicit} generates vector results from implicit signed distance fields (SDF) through a 2D Marching cube algorithm, leading to free-form contours. Therefore the edges are not smooth enough and have unpleasing noises, such as the `g' and `x' in the figure. Attr2Font\cite{attribute2font} can generate generally satisfactory font images, but it cannot maintain high-quality rendering when scaling. 

\begin{table}%
\caption{Quantitative comparison of image-level metrics of Attr2Font\cite{attribute2font}, Im2Vec\cite{im2vec}, Multi-Implicits\cite{multi-implicit}, DeepVecFont\cite{deepvecfont} and \sysname in the font complement experiment.}
\label{tab:complement}
\begin{minipage}{\columnwidth}
\begin{center}
\begin{tabular}{rcccc}
  \toprule
  Method & Resolution & SSIM$\uparrow$ & L1$\downarrow$ & s-IoU$\uparrow$ \\
  \midrule
  \multirow{3}*{\textbf{Ours}} & $128^2$ & \tb{0.8288}  & \tb{0.0461} & \tb{0.7395} \\
  & $256^2$ &  \tb{0.8727} & \tb{0.0460} & \tb{0.7453} \\
  & $512^2$ &  \tb{0.9075} &  \tb{0.0460} & \tb{0.7469} \\
  \midrule
  
  \multirow{3}*{DeepVecFont} & $128^2$ & 0.8165 & 0.0506 & 0.7125 \\
  & $256^2$ & 0.8640 & 0.0506 & 0.7171\\
  & $512^2$ & 0.9011 & 0.0506 & 0.7182\\
  \midrule
  
  \multirow{3}*{Multi-Implicits} & $128^2$ & 0.8025 & 0.0578 & 0.6857 \\
  & $256^2$ & 0.8533 & 0.0603 & 0.6814 \\
  & $512^2$ & 0.8914  & 0.0608 &  0.6811 \\
  \midrule
  
  \multirow{3}*{Im2Vec} & $128^2$  & 0.7576 & 0.0733 & 0.5731 \\
  & $256^2$ & 0.8270 & 0.0734 & 0.5751 \\
  & $512^2$ & 0.8739 & 0.0734 & 0.5756 \\
  \midrule
  
  Attr2Font\footnote{\cite{attribute2font} synthesizes pixelated images only. } & $128^2$ & 0.7841 & 0.0512 & 0.6393 \\
  \bottomrule
\end{tabular}
\end{center}
\centering
\end{minipage}
\vspace{-8mm}
\end{table}

\begin{table}%
\caption{The average number of commands used per glyph for different methods in the font generation task. M, L, Q, and C denote move, straight line, quadratic B\'ezier curve, and cubic B\'ezier curve respectively.}
\label{tab:command}
\centering
\scalebox{0.9}{
\begin{tabular}{rcccc|c}
  \toprule
   Command & M & L & Q & C & Total$\downarrow$ \\
  \midrule
   \textbf{Ours} & 1.42 & 8.21 & 11.06 & 0 & 20.69 \\
   DeepVecFont~\cite{deepvecfont} & 1.37 & 8.23 & 0 & 9.68 & \tb{19.28} \\
   Multi-Implicits~\cite{multi-implicit} & 6.16 & 1446.53 &  0 & 0 & 1452.69\\
  Im2Vec~\cite{im2vec} & 4 & 0 & 0 & 80 & 84 \\
  \midrule
   Human-Designed & 1.38 & 9.17 & 0 & 8.51 & 19.06 \\
  \bottomrule
\end{tabular}}



\vspace{-3mm}
\end{table}

Compared with the above methods, our method is capable of not only grasping the overall style of the input references but also generating new glyphs with clear boundaries and sharp edges, achieving optimal visual quality. The efficient dual-part representation along with the differentiable occupancy supervision establishes a link between pixelated images and vector graphics, allowing the unsupervised synthesis of vector fonts with aligned shapes and correct topologies. The refinement step allows the composition of the contour to be freely changed and thus ensures that our synthesized glyphs have high-fidelity details under the guidance of high-resolution glyph images. 
Quantitative results on three image-level metrics at different resolutions are shown in \cref{tab:complement}. For methods producing vector shapes, we also count the average number of commands used for each synthesized vector character, with smaller numbers indicating more compact outputs. The results are shown in \cref{tab:command}. Multi-Implicits\cite{multi-implicit} uses a large number of short line segments to represent the boundary due to the marching cube algorithm. Im2Vec\cite{im2vec} represents a shape with 4 parts each enclosed by 20 cubic B\'ezier curves, lacking flexibility and details. Our method performs boolean operations on the quadratic B\'ezier curves from dual parts, achieving comparable compactness with human-designed fonts and outputs from methods with vector supervision~\cite{deepvecfont}. 

\subsection{Ablation Studies}
\begin{figure}[th]
    \vspace{-5mm}
    \centering
    \setlength{\belowcaptionskip}{-10pt}
    \scalebox{0.85}{
    \begin{overpic}[scale=0.475]{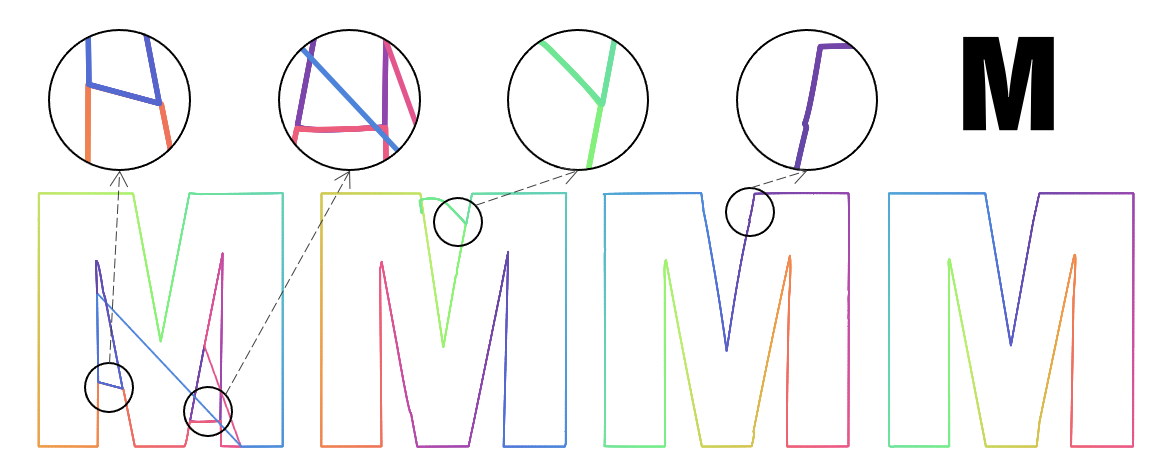}
    \put(32,-8){(a)}
    \put(100,-8){(b)}
    \put(168,-8){(c)}
    \put(218,-8){\textbf{Full Model}}
    \put(234,68){GT}
    \end{overpic}}
    \caption{The contours of an ``M" reconstructed by our method, compared with three degraded variants.}
    \label{fig:ablation}
\end{figure}
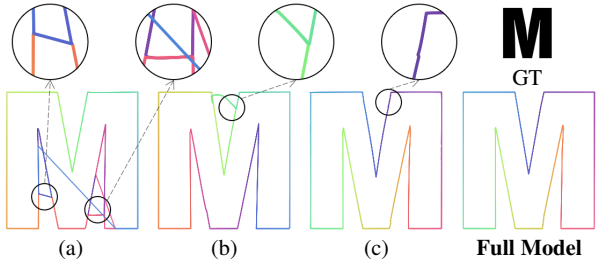

To demonstrate the effectiveness of several key design choices in our method, we conduct ablation studies on the font reconstruction task. We investigate the following degraded variants: (a) replace the union of dual parts with the union of all the paths; (b) train without the UDF warm-up process; and (c) remove the refining strategy during refinement. We show the comparison results of a representative case in \cref{fig:ablation}. Without the dual-part representation, variant (a) cannot reconstruct concave regions very well, as they can be hard to represent by the union of simply closed paths. Without UDF initialization, variant (b) produces contours with useless curves that are hard to be eliminated during optimization. Without the refining strategy, variant (c) may generate inaccurate edges with zigzags. Quantitative results in \cref{tab:ablation} also verify the quality improvements brought by the three design choices.

\begin{table}[t]%
\caption{Quantitative results of reconstruction accuracy for different model variants in ablation studies. All the output vector glyphs are rendered at $512\times512$ to compute the L1 error.}
\label{tab:ablation}
\centering
\begin{tabular}{l|ccc}
  \toprule
   & L1$\downarrow$ & $\#$Lines$\downarrow$ & $\#$ Curves$\downarrow$ \\
  \midrule
  (a) w/o dual parts & 0.0143 & 12.23 & 14.06 \\
  (b) w/o warm-up & 0.0146 & 11.19 & 13.12 \\
  (c) w/o refining strategy & 0.0148 & - & 22.07 \\
  \midrule
  
\tb{Full Model} & \tb{0.0136} & \tb{8.97} & \tb{11.81} \\
  \bottomrule
\end{tabular}
\vspace{-3mm}
\end{table}

We also experimentally investigate the effect of different $(N, M)$s on the representation capacity. We choose several settings of $(N, M)$ and train only the vector branch (i.e. no contour refinement) with the same number of epochs to reconstruct the input. 
\cref{tab:mn} shows that $M = 4$ is optimal and the reconstruction accuracy of the vector branch increases as $N$ increases. The accuracy does not gain too much changing $N$ from 6 to 8. So considering the time cost for training and inference, we select $(N, M) = (6, 4)$ in the above experiments. 

\vspace{-3mm}
\begin{table}[th]%
\caption{Different reconstruction settings and metrics. The image-based metrics use a resolution of $512\times512$. 
}
\vspace{-2mm}
\label{tab:mn}
\centering
\begin{tabular}{ccccc}
  \toprule
  $N$ & $M$ & SSIM$\uparrow$ & L1$\downarrow$ & s-IOU$\uparrow$\\ 
  \midrule
  2 & 4 & 0.9262 & 0.0282 & 0.8338 \\ 
  4 & 4 & 0.9292 & 0.0259 & 0.8464 \\
  6 & 4 & 0.9303 & 0.0250 & 0.8512 \\
  8 & 4 & 0.9306 & 0.0247 & 0.8525 \\
  6 & 3 & 0.9293 & 0.0256 & 0.8480 \\
  6 & 5 & 0.9300 & 0.0252 & 0.8498 \\
  \bottomrule
\end{tabular}
\vspace{-3mm}
\end{table}


\subsection{Font Sampling and Interpolation}
\begin{figure}[htbp]
\vspace{-3mm}
\setlength{\belowcaptionskip}{-15pt}
\includegraphics[width=\linewidth]{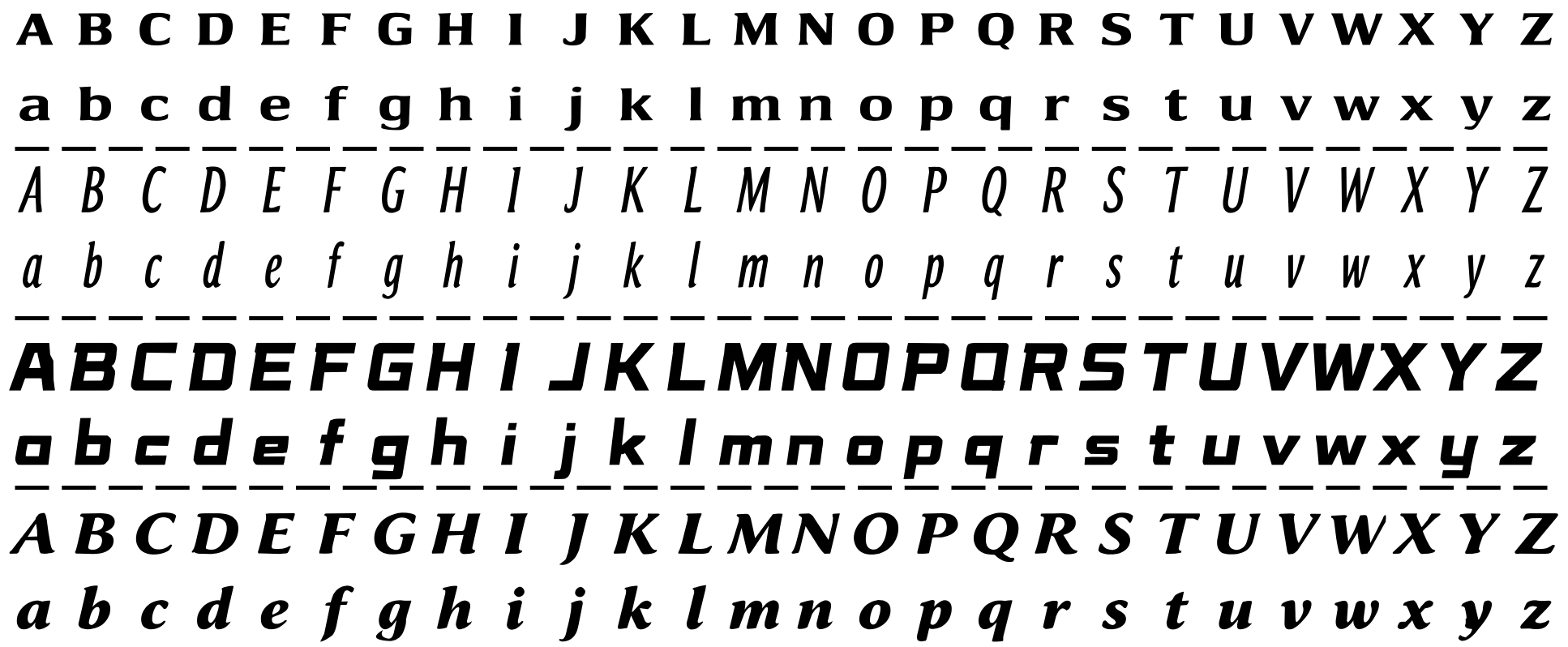}
\caption{New fonts with style codes sampled in $\mathcal{N}(0,1)$.}
\label{fig:sample}
\end{figure}

In the font generation task, the style latent code $z_{style}$ is trained with a variational loss $\mathcal{L}_{kl}$, enabling generating fonts of new styles by sampling $z_{style}$ from $\mathcal{N}(0,1)$.
We show in \cref{fig:teaser} (left) and \cref{fig:sample} multiple styles of fonts sampled.
Please refer to our supplementary material for more examples. 
Also, benefiting from our dual-part representation, we can perform smooth interpolation between arbitrary font styles as demonstrated in \cref{fig:teaser} (right).

\section{Limitations}
The contour refinement step requires gradient descent during inference and therefore imposes a large time overhead, similar to DeepVecFont\cite{deepvecfont}.
The process could potentially be accomplished by forwarding inferences of some generation models, such as the diffusion models\cite{ddpm}. Another limitation is that \sysname only focuses on the synthesis of glyphs on a fixed-size canvas, without considering the kerning between them, making the spacing between characters less natural. This may be solved by post-processing the SVG with some automatic kerning methods or by learning through a data-driven approach. 


\section{Conclusion}
We present \sysname as a new representation and generation method for vector fonts. \sysname models a glyph as dual parts and introduces a contour refinement step to add pleasing details to the contours of the dual parts. Our method is able to generate vector fonts without vector supervision, which can be directly converted to common digital font formats for content presentation and distribution. Comparisons with different methods of font reconstruction and generation tasks demonstrate that our approach produces fonts with the best quality among all the alternatives. In the future, more plausible initialization and supervision could be exploited to generalize this representation to more complex fonts (e.g. Chinese fonts). We hope this research contributes to the typeface design industry and provide inspiration for broader works on the understanding and generation of 2D man-made content such as cartoon characters and icons.

\noindent\textbf{Acknowledgement} This work was supported by the Natural Science Foundation of China (Project Number 61832016), and Tsinghua-Tencent Joint Laboratory for Internet Innovation Technology.

{\small
\bibliographystyle{ieee_fullname}
\bibliography{egbib}
}

\end{document}



\title{
\sysname: Unsupervised Vector Font Synthesis with Dual-Part Representation \\ Supplementary Material}

\author{
Ying-Tian Liu$^{1}$\quad
Zhifei Zhang$^{2}$\quad
Yuan-Chen Guo$^{1}$\quad
Matthew Fisher$^{2}$\\
Zhaowen Wang$^{2}$\quad
Song-Hai Zhang$^{1*}$\\
$^{1}$ BNRist, Department of Computer Science and Technology, Tsinghua University\;
$^{2}$ Adobe Research\\
{\tt\small \{liuyingt20@mails., guoyc19@mails.,  shz@\}tsinghua.edu.cn}\\
{\tt\small\{zzhang, matfishe, zhawang\}@adobe.com}
}



\maketitle


\appendix
\section{Implementation Details}

\newcommand{\lossc}[1]{\textcolor[RGB]{31,180,92}{#1}}
\newcommand{\udfc}[1]{\textcolor[RGB]{180,92,31}{#1}}
\subsection{Network Architecture and Evaluation}
In our implementation of \sysname, we set the number of dual parts and curves as $N = 6$ and $M = 4$.  The architecture of the image encoder $\mathcal{E}, \mathcal{E}_f$ and the decoder $\mathcal{D}_I$ is shown in \cref{tab:arch}, which is similar architecture to that presented in VQGAN~\cite{vqgan}. The output image $\mathcal{I}$ is with a resolution of $256\times256$. The path decoder $\mathcal{D}_P$ is implemented as an MLP in the size of $[256,256,256,8MN]$. The dimension for the intermediate variables $z, f_i, \mu, \sigma, T$ is set to $256$. In the font generation task, SA contains 4 stacked self-attention layers each with 4 attention heads and the FFN contains a linear layer in the size of $[512,256]$. We train the model for font reconstruction using Adam optimizer~\cite{adam} with a learning rate of $1\times10^{-3}$ with cosine annealing decay~\cite{cosannealing}. For the fine-tuning and refinement step, we adopt a learning rate of $2.5\times10^{-4}, 0.5$, respectively. In the contour refinement step, the canvas of the SVG glyph is set to have a side length of 256. For the font reconstruction task, we set $\lambda$'s as $\lambda_P = 0.5, \lambda_I = 1, \lambda_{u} = 1$ and the UDF warm-up is adopted in the first 8 epochs, lasting $\sim160$ epochs in all. For the font generation task, we first train the encoding parts, i.e. $\mathcal{E}_f$, SA, FFN and the embedding producing $T$, with the latent guidance and KL loss for 32 epochs to achieve a good initialization. We set $\lambda_{latent} = 20, \lambda_{kl} = 1.25\times 10^{-4}$ during this process. After that, we unfreeze the whole network and train it end-to-end with $\lambda_{P} = 1, \lambda_{I} = 0.5, \lambda_{kl} = 1.25\times 10^{-4}$ until convergence. Due to the limited speed of DeepVecFont~\cite{deepvecfont} and our method in generating fonts, we only evaluate the metrics on the first 200 fonts of the 1425 fonts in our all experiments for all methods.

\begin{table*}%
\caption{The architecture of the image encoder $\mathcal{E}, \mathcal{E}_f$ and the decoder $\mathcal{D}_I$, which is similar to that presented in \cite{vqgan}. The non-linear activation, Swish~\cite{swish}, is omitted. The input shape to the image encoder is $1 \times 128 \times 128$.} 
\label{tab:arch}
\centering
\begin{tabular}{c|c|c}
  \toprule
  Module & Block/Layer & Output Shape \\
  \midrule
  \multirow{9}*{Encoder $\mathcal{E}, \mathcal{E}_f$} & Conv2D & $16 \times 128 \times 128$\\
  & Residual Block + Downsample Block & $16 \times 64 \times 64$ \\
   & Residual Block + Downsample Block & $32 \times 32 \times 32$ \\
  & Residual Block + Attention Block + Downsample Block & $32 \times 16 \times 16$ \\
  & Residual Block + Downsample Block & $64 \times 8 \times 8$ \\
 & Residual Block + Downsample Block & $64 \times 4 \times 4$ \\
  & Residual Block + Downsample Block & $128 \times 2 \times 2$ \\
  & Residual Block + Downsample Block & $256 \times 1 \times 1$ \\
  & Residual Block & $256 \times 1 \times 1$ \\
  \midrule
  \multirow{11}*{Decoder $\mathcal{D}_I$}& Conv2D & $256 \times 1 \times 1$\\
  & Residual Block + Upsample Block & $256 \times 2 \times 2$ \\
  & Residual Block + Upsample Block & $128 \times 4 \times 4$ \\
  & Residual Block + Upsample Block & $128 \times 8 \times 8$ \\
  & Residual Block + Upsample Block & $64 \times 16 \times 16$ \\
  & Residual Block + Upsample Block & $64 \times 32 \times 32$ \\
  & Residual Block + Upsample Block & $32 \times 64 \times 64$ \\
  & Residual Block + Attention Block + Upsample Block & $32 \times 128 \times 128 $ \\
  & Residual Block + Upsample Block & $16 \times 256 \times 256 $ \\
  & Residual Block & $16 \times 256 \times 256 $ \\
  & GroupNorm + Conv2D & $1 \times 256 \times 256 $ \\
  \bottomrule
\end{tabular}
\end{table*}

\subsection{Contour Refinement}
To be clear, the contour $\partial{\mathcal{O}}$ contains multiple closed paths. Each path may consist of a different number of segments that are either straight line segments or B\'ezier curve segments. The SVG canvas is of size $256\times256$ in this stage. The contour refinement for each glyph lasts for 200 steps. We set $\lambda_{reg} = 10^{-6}$ for refinement. 
\paragraph{Pruning} Prior to the refinement, we first prune the paths $\{C_1, C_2, ..., C_K\}$ on the contour. Let $S(\cdot)$ denote the area. We remove $C_i$ that satisfies $S(C_i) < 50$.

\paragraph{Simplification} There are three different simplification strategies. (a) For every 50 iteration steps, we replace segments that are too short (with length $< 3$) with the average of their start and end points. (b) After that, we replace the B\'ezier curves $(\bm{a}, \bm{b}, \bm{c})$ with $\vec{\bm{ac}}$ if $\angle\bm{abc} > 171^{\circ}$. (c) After 150 iterations, we join the adjacent segments if they could be replaced by a single segment without losing many details. To be concrete, we join two adjacent line segments if they are at an angle greater than $175^{\circ}$. For adjacent B\'ezier curves, we first convert them to the implicit quadratic curve form. The differences between the coefficients of the implicit curves can roughly reflect their proximity. If $||e_1 - e_2||_2$ is less than 0.02 where $e_1, e_2$ are their normalized coefficients respectively, we consider them close enough and join them.

\paragraph{Subdivision} After 50 iterations, we split B\'ezier curves with a length greater than 25.6 (10\% of the canvas width) at their midpoints. 

\begin{figure*}[htbp]
    \centering
    \includegraphics[width=\linewidth]{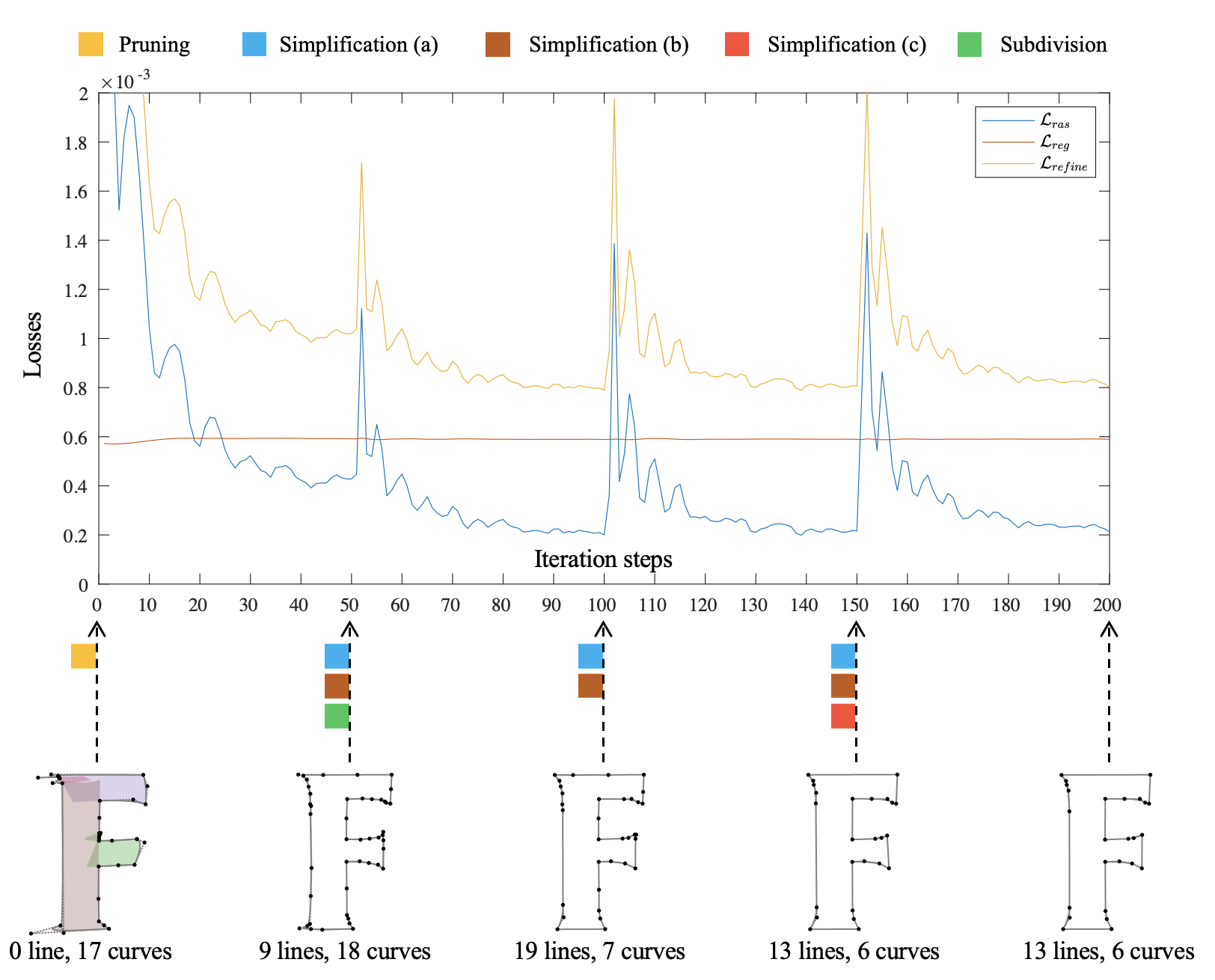}
    \caption{An example of the refinement process. We plot the loss $\mathcal{L}_{refine}$ as a function of iteration steps. The operations are marked near the timeline with different colors. We also show the contour along with the endpoints and control points throughout the refinement. We record the number of lines and curves of the contour under each intermediate result to illustrate how the composition of the contour changes as the refinement goes on.}
    \label{fig:refine}
\end{figure*}

In \cref{fig:refine}, we show an example of the refinement process, including how the loss changes with the iteration step, the strategy adopted at certain times, and the intermediate results. The figure illustrates that our contour refinement step can improve the quality and representation efficiency of the contours.

\section{Unsigned Distance Field Warm-Up}
\begin{figure*}[htbp]
\centering
\begin{overpic}[scale=0.4]{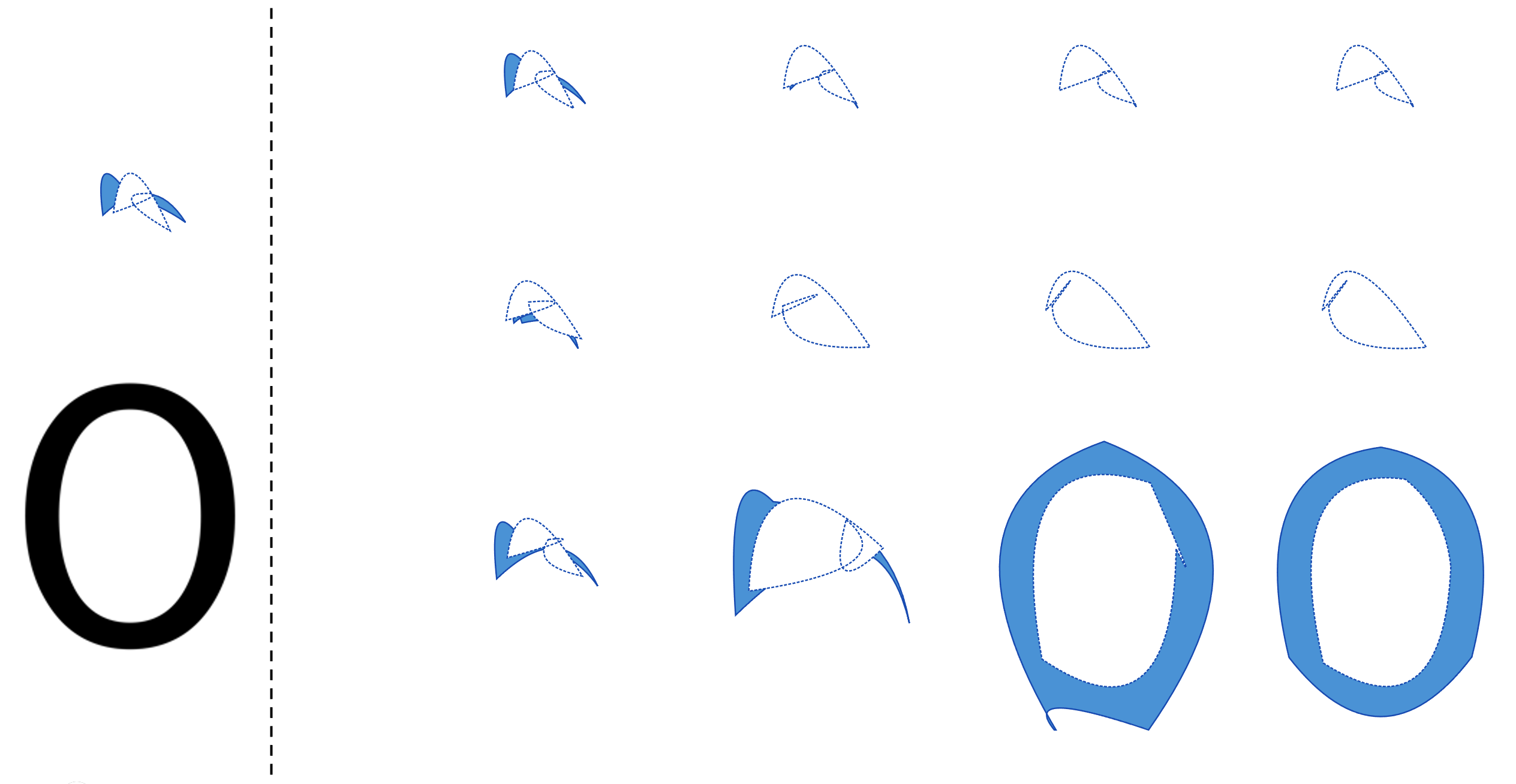}
\put(35,170){\Large Init}
\put(27,25){\Large Target}
\put(115,230){DiffVG~\cite{diffvg}}
\put(125,150){Ours $\mathcal{L}_{P}$}
\put(105,70){Ours $\mathcal{L}_{P}$+$\mathcal{L}_{u}$}
\put(170,260){Iter 1}
\put(262,260){Iter 5}
\put(354,260){Iter 50}
\put(446,260){Iter 200}

\put(170,200){\lossc{0.152}}
\put(263,200){\lossc{0.148}}
\put(356,200){\lossc{0.148}}
\put(450,200){\lossc{0.148}}

\put(170,120){\lossc{0.149}}
\put(263,120){\lossc{0.148}}
\put(356,120){\lossc{0.148}}
\put(450,120){\lossc{0.148}}

\put(170,10){\lossc{0.154}}
\put(263,10){\lossc{0.127}}
\put(356,10){\lossc{0.018}}
\put(450,10){\lossc{0.009}}

\put(170,-2){\udfc{0.344}}
\put(263,-2){\udfc{0.185}}
\put(356,-2){\udfc{0.012}}
\put(450,-2){\udfc{0.007}}
\end{overpic}
\caption{A comparison between the pixel loss of DiffVG~\cite{diffvg}, our occupancy loss in Eq.(10) in the main paper, and our occupancy loss with UDF warm-up. We optimize a randomly initialized dual part with 6 B\'ezier curves in the positive and negative parts with the above three losses. We plot the dual part at the 1st, 5th, 50th, and 200th iteration steps. The positive path is solid and the negative one is drawn with dashed lines. We also mark \lossc{the mean square error} between the rendered image and the target and \udfc{the UDF warm-up loss $\mathcal{L}_u$} under each dual part.}
\label{fig:optimize}
\end{figure*}

In \cref{fig:optimize}, we illustrate how the UDF warm-up strategy could help the optimization of the dual part parameters. In this case, we optimize a randomly initialized dual part to a letter `O' using different losses. The figure shows that using the pixel loss between the rendered image obtained by DiffVG~\cite{diffvg} and the target or the loss $\mathcal{L}_{P}$ based on a differentiable occupancy field will yield similar results. This is because both losses can only produce gradients near the edges of the shape. When shapes fall into a hole, the losses tend to shrink the positive path and expand the negative path until they produce an entirely white solid shape. The UDF loss $\mathcal{L}_u$ can drive the shape to the approximately correct position so that it does not fall into local minima, producing results with minimal error. Therefore, we claim that the UDF warm-up will help with the initialization of dual parts in the early stage of training.

\section{Font Sampling and Interpolation Results}
We show more font sampling results in \cref{fig:sample_supp} in SVG format. 
Also, we provide font interpolation results in \cref{fig:inter_supp}. The fonts at both ends of the interpolation are also generated by \sysname. We obtain intermediate results by performing linear interpolation in the latent space of the font style codes. We also provide in the attachment files some TrueType fonts which can be installed on the computer to display in various text editors. They are converted from \sysname's SVG results.

\section{Metric for Vector Fonts}
 Evaluating the quality of synthetic fonts based on metrics in the vector domain is valuable. But designing a reasonable metric on vector graphics remains an open problem and is worth investigating. Here we devise a Chamfer-Distance-style metric between two vector glyphs. Suppose the two glyphs have contours $U, V$ on the unit square canvas, we uniformly sample $n$ points $\{u_i\}_{i=1}^n$ on $U$ and calculate their average distance to V and vice versa. The ``vector distance'' between $U, V$ is defined as
\begin{equation*}
    d_{VD}(U, V) = \frac 1 n \sum_{i=1}^n\min_{v\in V} ||u_i-v||_2 + \frac 1 n \sum_{i=1}^n\min_{u\in U} ||v_i-u||_2   
\end{equation*}
We evaluate the metric ($n = 100$) for both font reconstruction and generation tasks (\cref{tab:vd_recon}). \sysname achieves the best fit in terms of contouring. 
\vspace{-3mm}
\begin{table}[h]%
\caption{Evaluation of the ``vector distance'' metric for the font reconstruction and generation tasks.}
\vspace{-2mm}
\label{tab:vd_recon}
\centering
\begin{tabular}{rcc}
  \toprule
  Method & $d_{VD}$(recon)$\downarrow$ & $d_{VD}$(generation)$\downarrow$ \\
  \midrule
  \textbf{Ours} & \textbf{0.0082} & \textbf{0.0346} \\
  DeepVecFont & - & 0.0491 \\
  Multi-Implicits & 0.0170 & 0.0395 \\
  Im2Vec & 0.0473 & 0.0658 \\
  \bottomrule
\end{tabular}
\vspace{-3mm}
\end{table}

\section{Compact and Topologically Correct Vector Fonts}

\begin{figure}[t]
\centering
\begin{overpic}[scale=0.45]{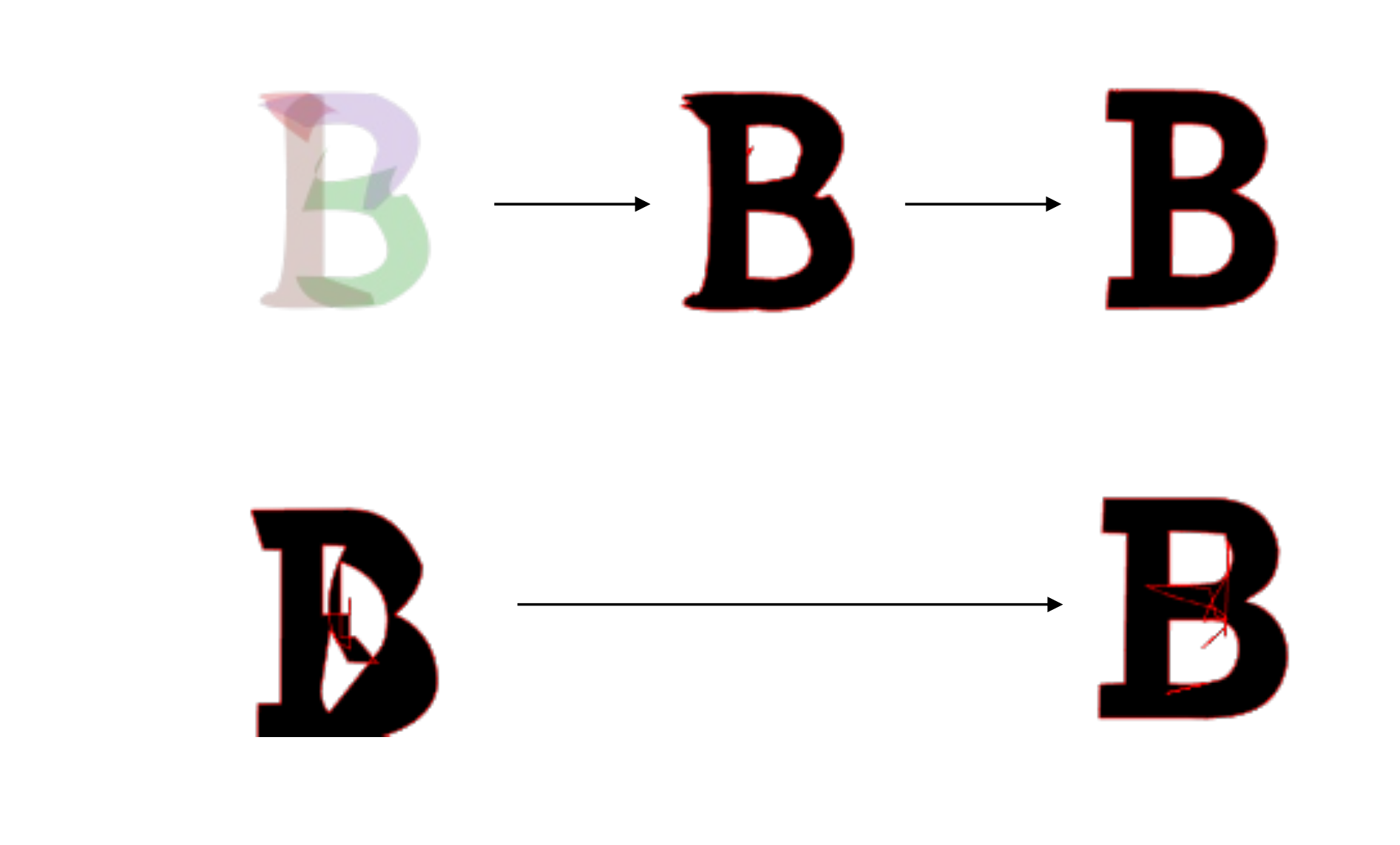}
\put(20,110){\textbf{Ours}}
\put(0,37){\scalebox{0.8}{DeepVecFont}}
\put(40,80){Dual Parts}
\put(110,80){Initial $\partial\mathcal{O}$}
\put(180,80){Vector Glyph}
\put(37,8){Initial SVG}
\put(180,8){Vector Glyph}
\end{overpic}

\caption{A comparison between our method and DeepVecFont~\cite{deepvecfont} in the process of synthesizing vector fonts. We produce dual parts and then calculate and simplify contours, while DeepVecFont~\cite{deepvecfont} predicts the contours directly and refines them later.}
\label{fig:compact}
\end{figure}

\sysname learns the shape decomposition from only glyph images and achieves compact vector font synthesis. But DeepVecFont~\cite{deepvecfont} may produce results with wrong topologies even with vector supervision. The difference is that DeepVecFont~\cite{deepvecfont} predicts the glyph contour directly, including the type and parameters of SVG commands, which are both predicted by the network. If the SVG command prediction is wrong, it will easily produce unwanted or self-intersecting curves. \sysname, on the other hand, separates shape prediction and SVG production, decreasing the difficulty. It outputs the parameters of dual parts and calculates the contour later. The contour calculation and refinement can avoid self-intersection and decide the type of SVG elements for each segment, respectively. We show a typical example in \cref{fig:compact}. DeepVecFont~\cite{deepvecfont} relies on the robustness of the network to ensure that the topology is reasonable. Our method eliminates the possibility of self-intersection by a deterministic contour calculation. In this way, \sysname produces a more compact vector glyph.

\section{Dual Part Correspondence}
In \cref{fig:parts}, we show the results of reconstructing several different styles of glyphs of the same letter with dual parts. We observe that the same dual part usually carries a single semantic meaning when representing the same letter in different styles, for example, the third  part (green) focuses on reconstructing the lower part of its right vertical line when representing `M'. The correspondence established by our method through dual parts is positive for downstream tasks such as interpolation or generation. To further illustrate the effect of such correspondence, we show the results of interpolating their dual parts between two styles of the same letter in \cref{fig:inter_A}. The correspondence between dual parts makes the interpolation very smooth, indicating the good properties of glyphs' latent space, which is why \sysname can generate high-quality glyphs.

\begin{figure*}[t]
    \centering
    \includesvg[inkscapelatex=false, width=0.9\linewidth]{0013_tune}
    \vspace{2mm}
    \rule[0pt]{\linewidth}{0.1em}
    \vspace{2mm}
    
    \includesvg[inkscapelatex=false, width=0.9\linewidth]{0016_tune}
     \vspace{2mm}
    \rule[0pt]{\linewidth}{0.1em}
    \vspace{2mm}
    \includesvg[inkscapelatex=false, width=0.9\linewidth]{0029_tune}
     \vspace{2mm}
    \rule[0pt]{\linewidth}{0.1em}
    \vspace{2mm}

    \includesvg[inkscapelatex=false, width=0.9\linewidth]{0047_tune}
 
    \caption{New fonts generated by \sysname in SVG format. }
    \label{fig:sample_supp}
\end{figure*}

\begin{figure*}[t]
    \centering
     \includesvg[inkscapelatex=false, width=0.8\linewidth]{0168_inter_00_52.svg} 
    
    \vspace{3mm}
    
     \includesvg[inkscapelatex=false, width=0.8\linewidth]{0168_0094_07_01_inter_00_52.svg} 
     
    \vspace{3mm}
    
    \includesvg[inkscapelatex=false, width=0.8\linewidth]{0168_0094_07_02_inter_00_52.svg}

    \vspace{3mm}
    
    \includesvg[inkscapelatex=false, width=0.8\linewidth]{0168_0094_07_03_inter_00_52.svg}
    
    \vspace{3mm}
    
    \includesvg[inkscapelatex=false, width=0.8\linewidth]{0168_0094_07_04_inter_00_52.svg}

    \vspace{3mm}
    
    \includesvg[inkscapelatex=false, width=0.8\linewidth]{0168_0094_07_05_inter_00_52.svg}

    \vspace{3mm}
    
    \includesvg[inkscapelatex=false, width=0.8\linewidth]{0094_inter_00_52.svg}

    \rule[0pt]{0.9\linewidth}{0.1em}

    \vspace{2mm}
   
    \includesvg[inkscapelatex=false, width=0.8\linewidth]{0067_inter_00_52.svg} 
    
    \vspace{3mm}
    
     \includesvg[inkscapelatex=false, width=0.8\linewidth]{0067_0034_07_01_inter_00_52.svg} 
     
    \vspace{3mm}
    
    \includesvg[inkscapelatex=false, width=0.8\linewidth]{0067_0034_07_02_inter_00_52.svg}

    \vspace{3mm}
    
    \includesvg[inkscapelatex=false, width=0.8\linewidth]{0067_0034_07_03_inter_00_52.svg}
    
    \vspace{3mm}
    
    \includesvg[inkscapelatex=false, width=0.8\linewidth]{0067_0034_07_04_inter_00_52.svg}

    \vspace{3mm}
    
    \includesvg[inkscapelatex=false, width=0.8\linewidth]{0067_0034_07_05_inter_00_52.svg}

    \vspace{3mm}
    
    \includesvg[inkscapelatex=false, width=0.8\linewidth]{0034_inter_00_52.svg}

    \caption{Font examples with linear interpolated styles in the latent space.}
    \label{fig:inter_supp}
\end{figure*}

\begin{figure*}[t]
    \centering
    \includegraphics[width=\linewidth]{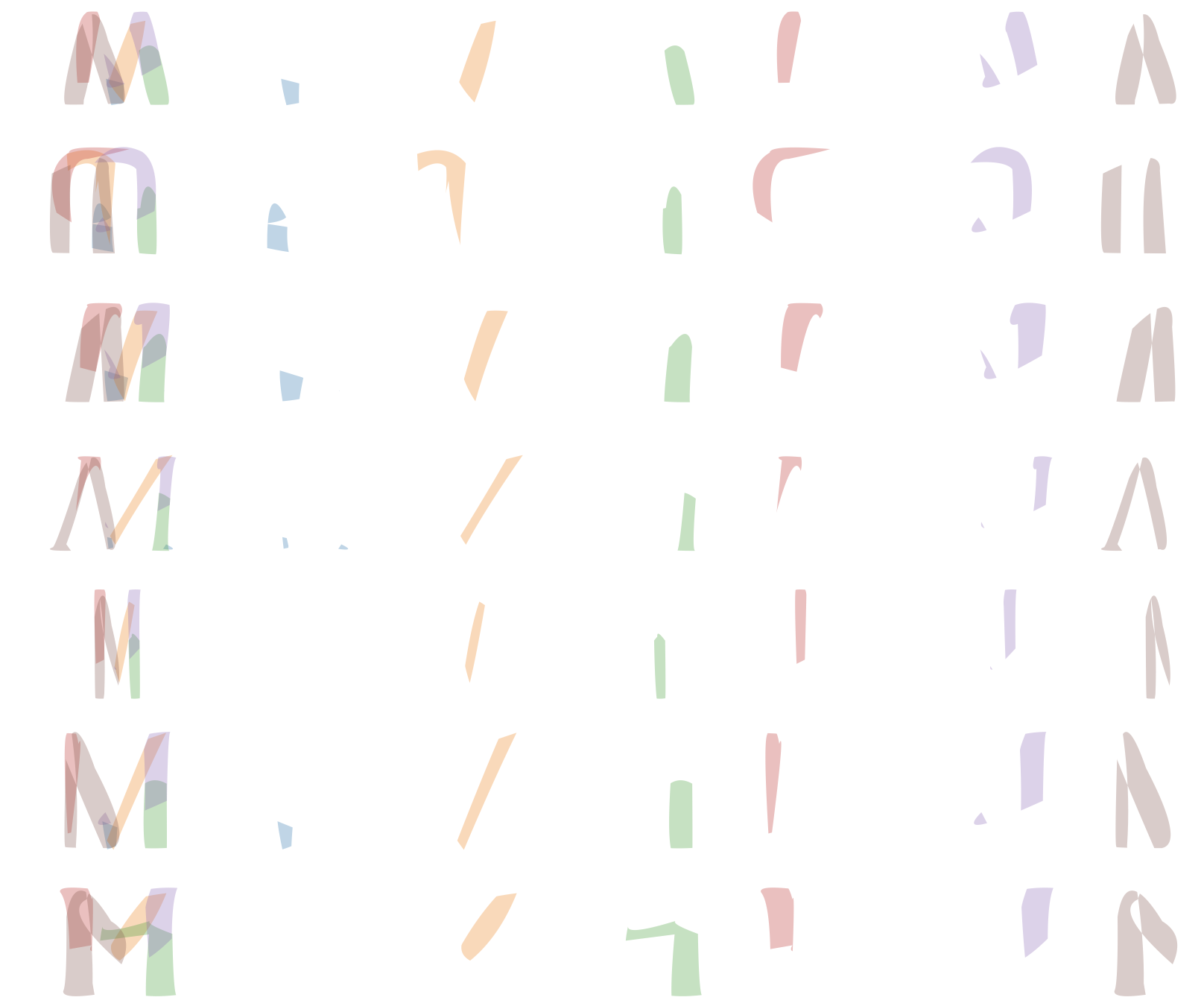}

    \caption{Different styles of the letter `M' are formed by dual parts with correspondence.}
    \label{fig:parts}
\end{figure*}
\begin{figure*}[t]
\ContinuedFloat
    \centering
    \includegraphics[width=\linewidth]{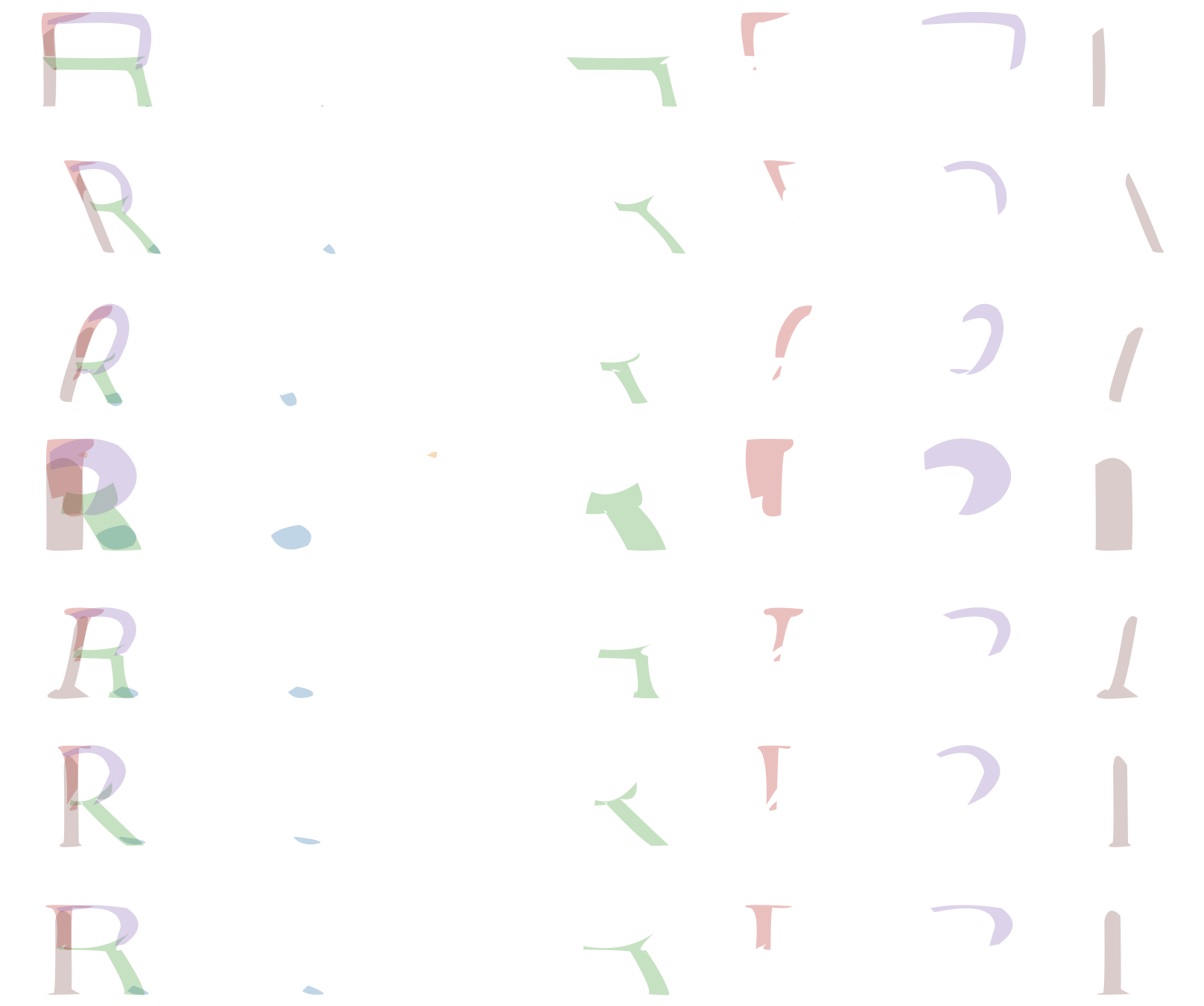}
    \caption{Different styles of the letter `R' are formed by dual parts with correspondence.}
\end{figure*}
\begin{figure*}[t]
    \ContinuedFloat
    \centering
    \includegraphics[width=\linewidth]{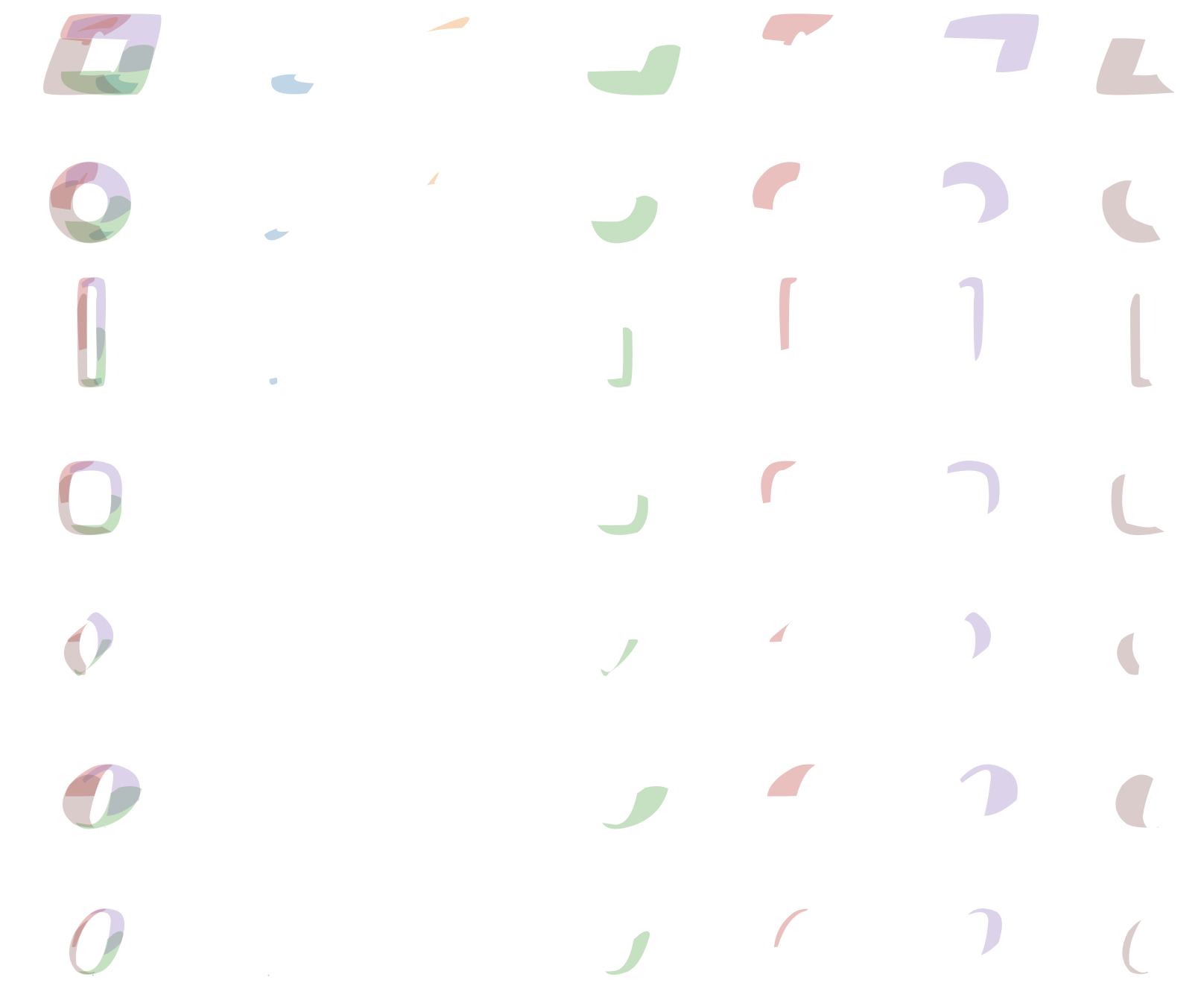}
    \caption{Different styles of the letter `o' are formed by dual parts with correspondence.}
\end{figure*}

\begin{figure*}[t]
    \centering
    \includegraphics[width=\linewidth]{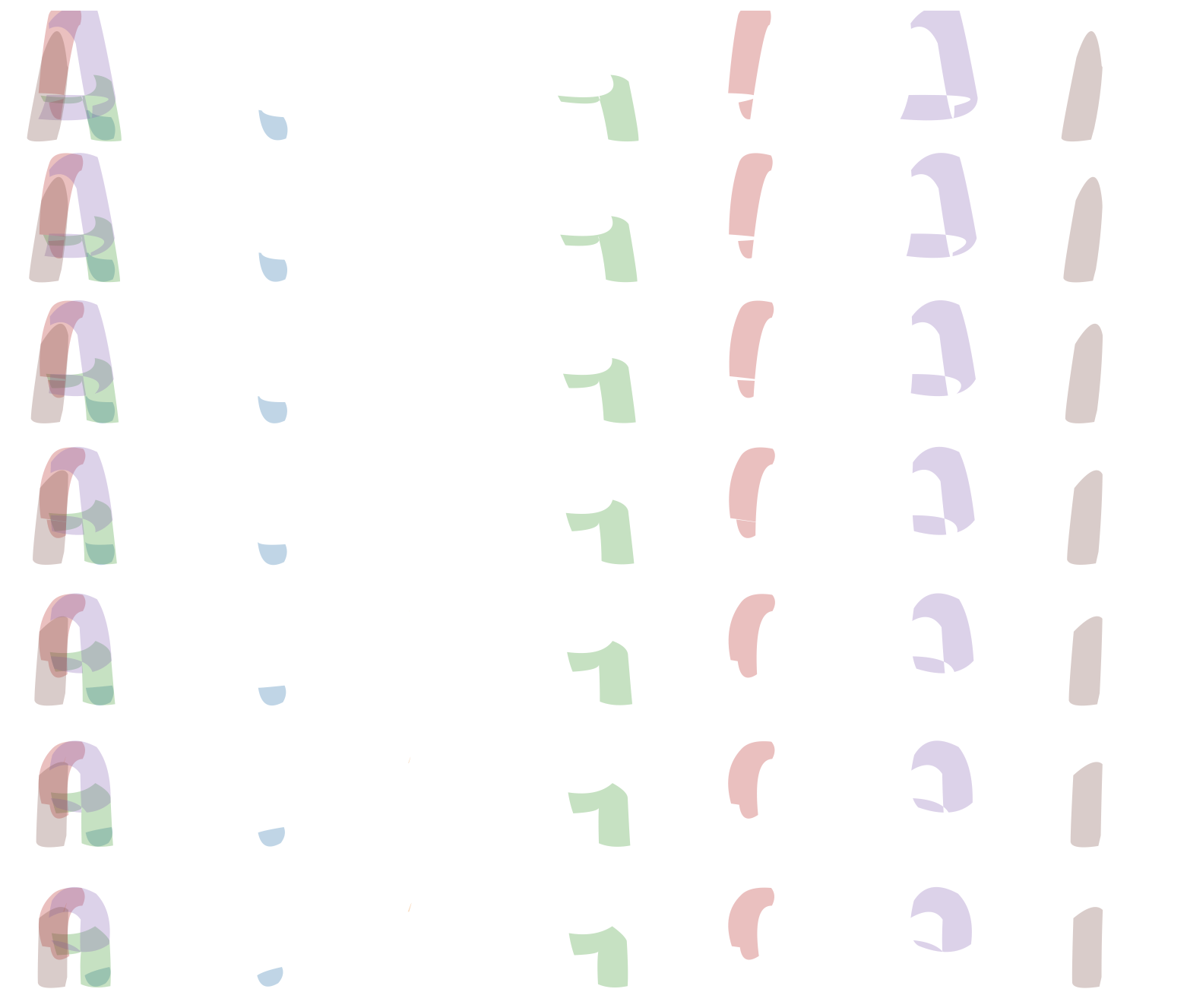}
    \caption{The interpolation of dual parts between two styles of `A'.}
    \label{fig:inter_A}
\end{figure*}
\begin{figure*}[t]
    \ContinuedFloat
    \centering
    \includegraphics[width=\linewidth]{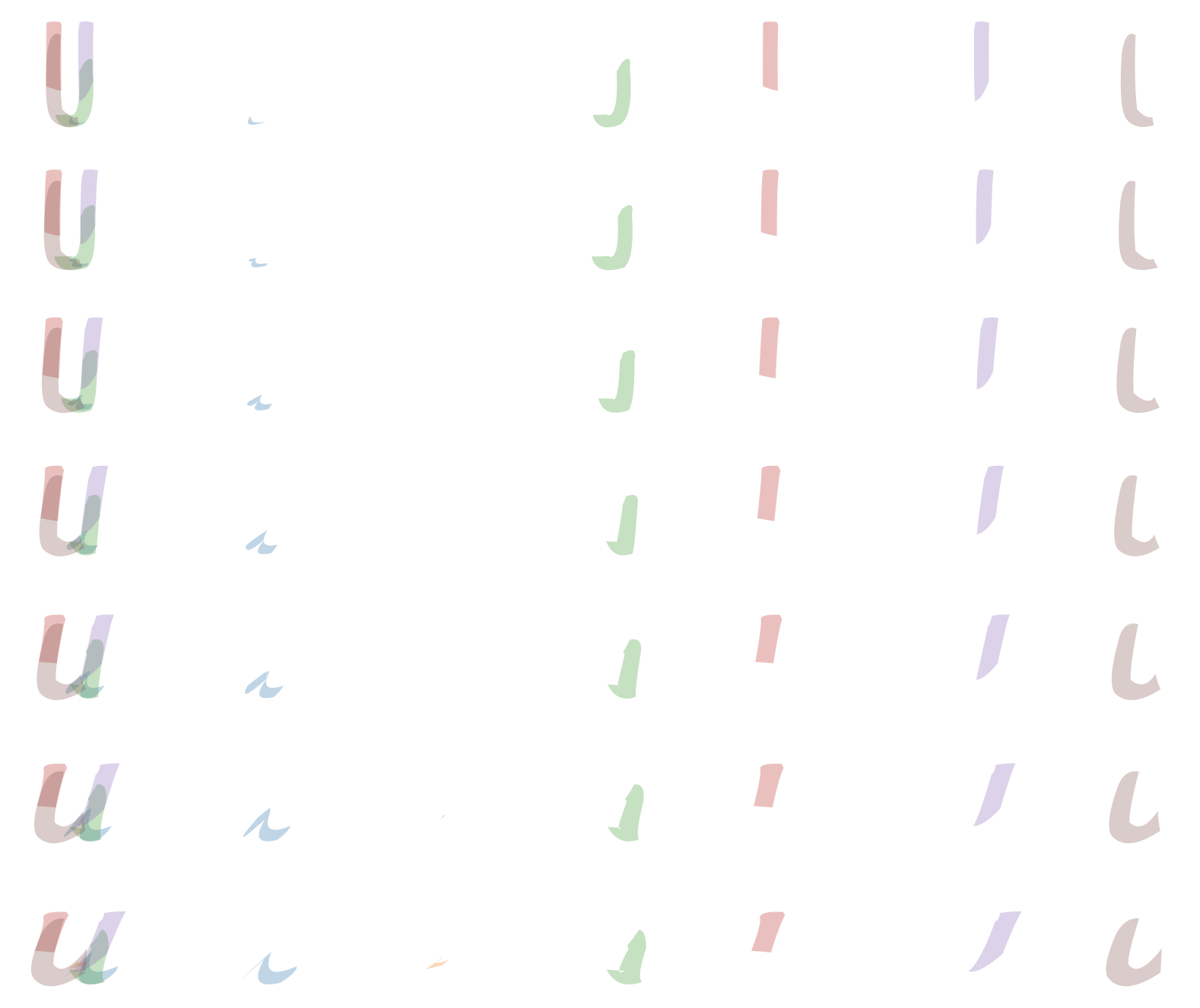}
    \caption{The interpolation of dual parts between two styles of `U'.}
\end{figure*}
{\small
\bibliographystyle{ieee_fullname}
\bibliography{egbib}
}